\documentclass{article}

\PassOptionsToPackage{numbers, compress}{natbib}

\usepackage[final]{neurips_2022}

\usepackage[utf8]{inputenc} %
\usepackage[T1]{fontenc}    %
\usepackage{hyperref}       %
\usepackage{url}            %
\usepackage{booktabs}       %
\usepackage{amsfonts}       %
\usepackage{nicefrac}       %
\usepackage{microtype}      %
\usepackage{xcolor}         %
\usepackage{listings}
\usepackage{algorithm}
\usepackage{algpseudocode}
\usepackage{wrapfig}

\usepackage{makecell}

\usepackage{amsthm}

\usepackage{amsmath,amsfonts,bm}

\def\eqref#1{equation~\ref{#1}}

\def\1{\bm{1}}

\def\rmd{\mathrm{d}}

\DeclareMathAlphabet{\mathsfit}{\encodingdefault}{\sfdefault}{m}{sl}
\SetMathAlphabet{\mathsfit}{bold}{\encodingdefault}{\sfdefault}{bx}{n}

\newcommand{\E}{\mathbb{E}}

\newcommand{\KL}{D_{\mathrm{KL}}}

\usepackage{cleveref}

\newtheorem{prop}{Proposition}

\newtheorem{assum}{Assumption}

\usepackage{hyperref}
\hypersetup{colorlinks,linkcolor={blue},citecolor={blue},urlcolor={red}}
\usepackage{graphicx}
\setcitestyle{square}

\title{Score-Based Diffusion  meets \\ Annealed Importance Sampling} %

\author{Arnaud Doucet, Will Grathwohl, Alexander G. D. G. Matthews \& Heiko Strathmann \thanks{alphabetical order, equal contribution.} \\
DeepMind\\
\texttt{\{arnauddoucet,wgrathwohl,alexmatthews,strathmann\}@google.com} \\
}

\begin{document}

\maketitle

\begin{abstract}
More than twenty years after its introduction, Annealed Importance Sampling (AIS) remains one of the most effective methods for marginal likelihood estimation. It relies on a sequence of distributions interpolating between a tractable initial distribution and the target distribution of interest which we simulate from approximately using a non-homogeneous Markov chain. To obtain an importance sampling estimate of the marginal likelihood, AIS introduces an extended target distribution to reweight the Markov chain proposal.
While much effort has been devoted to improving the proposal distribution used by AIS,
an underappreciated issue is that AIS uses a convenient but suboptimal extended target distribution. We here leverage recent progress in score-based generative modeling (SGM) to approximate the optimal extended target distribution minimizing the variance of the marginal likelihood estimate for AIS proposals corresponding to the discretization of Langevin and Hamiltonian dynamics. We demonstrate these novel, differentiable, AIS procedures on a number of synthetic benchmark distributions and variational auto-encoders.

\end{abstract}

\section{Introduction}
Evaluating the marginal likelihood, also known as evidence, is of key interest in Bayesian statistics as it allows not only model comparison but is also often used to select hyperparameters. A large variety of Monte Carlo methods have been proposed to address this problem, including path sampling \citep{gelman1998simulating}, AIS \citep{neal2001annealed} and related Sequential Monte Carlo methods \citep{Del-Moral:2006}. An appealing feature of AIS is that it provides an unbiased estimate of the marginal likelihood and can thus be used to define an evidence lower bound (ELBO) or mutual information bounds; see e.g. \citep{wunoe2020stochastic,thin2021MCVAE,brekelmans2021improving}.

AIS builds a proposal distribution using a Markov chain $(x_k)_{k=0}^K$ initialized at an easy-to-sample distribution followed by a sequence of Markov chain Monte Carlo (MCMC) transitions targeting typically annealed versions of the posterior. By proceeding this way, we obtain a proposal $x_K$ whose distribution is expected to be a reasonable approximation to the target posterior. %
However, this distribution is intractable as it requires integrating the joint proposal distribution over previous states $(x_k)_{k=0}^{K-1}$. AIS bypasses this issue by instead using Importance Sampling (IS) on the whole path $(x_k)_{k=0}^K$ through the introduction of an artificial extended target distribution whose marginal at time $K$ coincides with the posterior.

There has been much work devoted to improving AIS in machine learning and statistics but also in physics where it was introduced independently in \citep{jarzynski1997nonequilibrium,crooks1998nonequilibrium}. A standard approach to improve AIS is to modify the intermediate distributions  \citep{schmiedl2007optimal,Grosse2013,masrani2021q} and corresponding transition kernels of the proposal \citep{dai2020invitation,wunoe2020stochastic,Geffner:2021,thin2021MCVAE,Zhang2021}. We here address a distinct problem. For a given proposal, it was shown in \citep{Del-Moral:2006} that the extended target distribution minimizing the variance of the evidence estimate is not the one used by AIS but is instead defined through the time-reversal of the proposal.
However, this result is difficult to exploit algorithmically as the time-reversal is intractable for useful proposals.

\begin{figure}
    \centering
    \includegraphics[width=.99\textwidth]{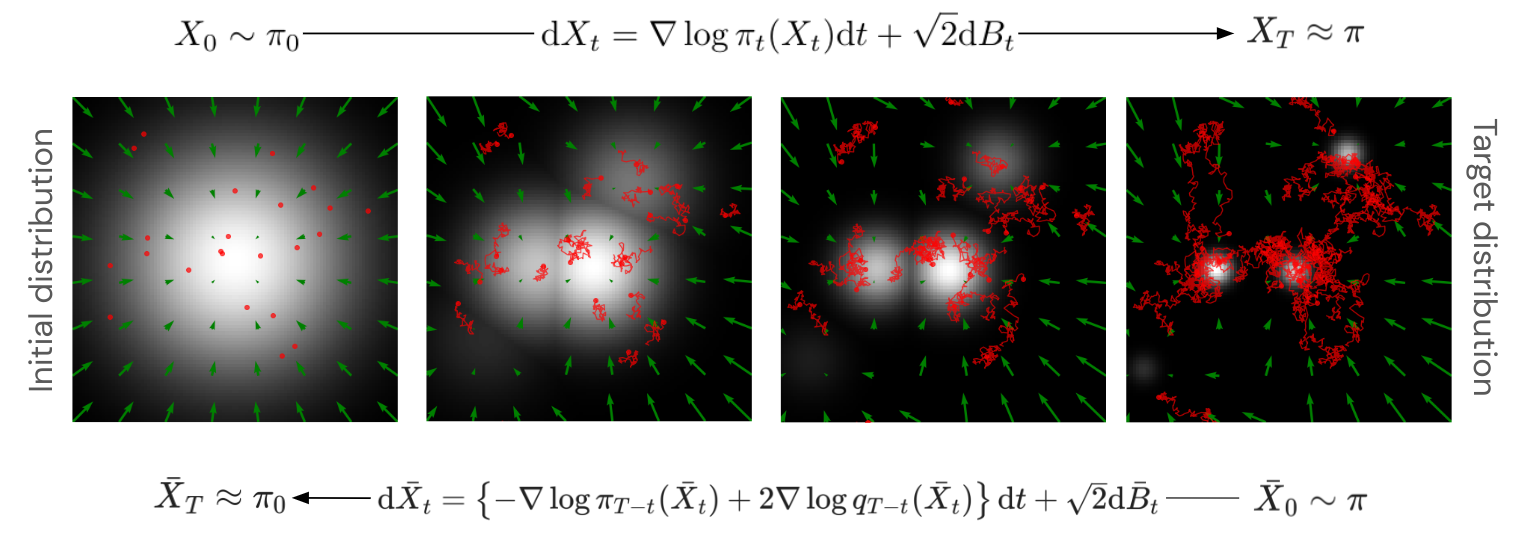}
    \caption{\textbf{Top}: Samples $X_t$ from an AIS proposal (red) obtained by sampling initially from a Gaussian at $t=0$ and diffusing through Langevin dynamics on intermediate targets $\pi_t$ (white). The intermediate marginals  of the proposal, $q_t$, approximated by the samples are such that $q_T \approx \pi$ for a reasonably fast mixing diffusion.
    \textbf{Bottom}: Computing importance weights. The optimal extended target used to compute the weights is the distribution obtained by initializing $\bar{X}_0$ exactly from $\pi$ and then following the reverse-time dynamics of the forward AIS proposal. This requires access to score vectors of the marginals $q_t$.}
    \label{fig:Langevin}
\end{figure}

In this paper, we show how one can combine this result with recent advances in SGM to obtain improved, lower variance, AIS estimates. We concentrate on scenarios where we use unadjusted overdamped Langevin \citep{heng2017controlled,wunoe2020stochastic,thin2021MCVAE} and unadjusted Hamiltonian proposals with partial momentum refreshment (i.e. underdamped Langevin) \citep{dai2020invitation,wunoe2020stochastic,Geffner:2021,Zhang2021,jankowiak2022surrogate} which correspond to time-discretized diffusion processes. The first benefit of using such proposals is that, by omitting Metropolis--Hastings steps, one obtains differentiable versions of the Evidence Lower Bound (ELBO) amenable to the reparameterization trick.~%
The second benefit of these proposals is that their time-reversal can be approximated by adapting techniques developed for SGM \citep{ho2020denoising,song2020score,Dockhorn2022} to our setup. We derive a principled parameterization for an approximation of their time-reversal which we learn by maximizing the ELBO. As for SGM, this ELBO coincides with a denoising score matching loss \citep{Hyvarinen:2005a,vincent2011connection,ho2020denoising,song2020score}. This provides novel, optimized and differentiable, AIS estimators which we refer to as Monte Carlo Diffusion (MCD). We demonstrate the benefits of this approach on synthetic benchmark distributions and variational auto-encoders (VAEs)~\citep{kingma2013auto}. All proofs can be found in the Appendix. A preliminary version of this work appeared in \citep{doucet2022annealed}.

\section{Annealed Importance Sampling}

\subsection{Setup and algorithm} Consider a probability density $\pi$ on  $\mathbb{R}^d$ of the form
\begin{equation}
    \pi(x)=\frac{\gamma(x)}{Z},\qquad Z=\int_{\mathbb{R}^d} \gamma(x)\rmd x,
\end{equation}
where $\gamma(x)$ can be evaluated pointwise. We want to approximate the intractable normalizing constant $Z$. In a Bayesian framework, $\gamma(x)=p(x)p(\mathcal{D}|x)$ is the joint density of parameter $x$ and data $\mathcal{D}$, $\pi(x)=p(x|\mathcal{D})$ the corresponding posterior and $Z=p(\mathcal{D})$ the evidence. 

To estimate $Z$, AIS introduces the intermediate distributions $(\pi_k)_{k=1}^K$ bridging smoothly from a tractable distribution $\pi_0$ to the target distribution $\pi_K=\pi$ of interest.
One typically uses  $\pi_k(x) \propto \gamma_k(x)$ with $\gamma_k(x)=\pi_0(x)^{1-\beta_k}\gamma(x)^{\beta_k}$ for $0=\beta_0<\beta_1<\cdots<\beta_K=1$ but other choices are possible~\citep{Grosse2013}.
The IS proposal used by AIS is then obtained by running a Markov chain $(x_k)_{k=0}^K$ such that $x_0\sim \pi_0(\cdot)$, and then $x_k\sim F_k(\cdot|x_{k-1})$ for $k\geq 1$ where $F_k$ is a MCMC kernel invariant w.r.t.\ $\pi_k$. The proposal is thus given by 
\begin{equation}\label{eq:AISproposal}
    \textstyle 
     Q(x_{0:K})=\pi_0(x_0) \prod_{k=1}^K F_{k}(x_k|x_{k-1}).
\end{equation}
Denote by $q_k$ the marginal distribution of $x_k$ under $Q$ satisfying $q_k(x_k)=\int q_{k-1}(x_{k-1})F_k(x_k|x_{k-1})\rmd x_{k-1}$ for $k\geq 1$ and $q_0=\pi_0$, it is typically intractable for $k\geq 1$. As $q_K$ cannot be evaluated in complex scenarios, the marginal IS estimate $w_{\textup{mar}}(x_K)=\gamma(x_K)/q_K(x_K)$ of $Z$ is intractable.

One can bypass this issue by introducing an extended target distribution
\begin{equation}\label{eq:AIStarget}
    P(x_{0:K})=\frac{\Gamma(x_{0:K})}{Z},\quad\quad \Gamma(x_{0:K})=\gamma(x_K)\prod_{k=0}^{K-1}B_k(x_k|x_{k+1}),
\end{equation}
where $(B_k)_{k=0}^{K-1}$ are backward Markov transition kernels, i.e. $\int B_k(x_k|x_{k+1})\mathrm{d}x_k=1$ for any $x_{k+1}$, so that by construction $x_K \sim \pi$ under $P$. For any selection of backward kernels such that the ratio $\Gamma/Q$ is well-defined, we then have 
\begin{equation}\label{eq:evidenceestimate}
   \mathbb{E}_Q[w(x_{0:K})]=Z,\quad \text{for~~}  w(x_{0:K})=\frac{\Gamma(x_{0:K})}{Q(x_{0:K})},
\end{equation}
i.e. $w(x_{0:K})$ is an unbiased estimate of $Z$ for $x_{0:K}\sim Q$.

The AIS estimate of the evidence is a specific instance of the estimator (\ref{eq:evidenceestimate}) relying on the backward kernels $B^{\textup{ais}}_{k}(x_k|x_{k+1})=\pi_{k+1}(x_{k})F_{k+1}(x_{k+1}|x_{k})/\pi_{k+1}(x_{k+1})$. This yields the following expression for $\log w(x_{0:K})$:
\begin{equation}\label{eq:logAISestimate}
\textstyle
     \log w_{\textup{ais}}(x_{0:K})= \sum_{k=1}^K \log \big(\gamma_k(x_{k-1})/\gamma_{k-1}(x_{k-1})\big). 
\end{equation}

\subsection{Limitations of AIS} While designing $P$ in (\ref{eq:AIStarget}) by using the backward Markov kernels $(B^{\textup{ais}}_{k})_{k=0}^{K-1}$ is convenient, it is also suboptimal in terms of variance. For example, consider the ideal scenario where $F_k(x_k|x_{k-1})=\pi_k(x_k)$. This scenario has been used many times in the literature to provide some guidelines on AIS, see e.g. \cite{neal2001annealed,Grosse2013}. 
In this case, $\textup{var}_Q[\log w_{\textup{ais}}(x_{0:K})]=\sum_{k=1}^K \textup{var}_{\pi_{k-1}}[\log (\gamma_k(x_{k-1})/\gamma_{k-1}(x_{k-1}))]>0$ while  $\textup{var}_{q_K}[w_{\textup{mar}}(x_{K})]=\textup{var}_{\pi}[w_{\textup{mar}}(x_{K})]=0$.

\begin{wrapfigure}{r}{0.4\textwidth}
\vspace{-0.7cm}
  \begin{center}
    \includegraphics[width=0.38\textwidth]{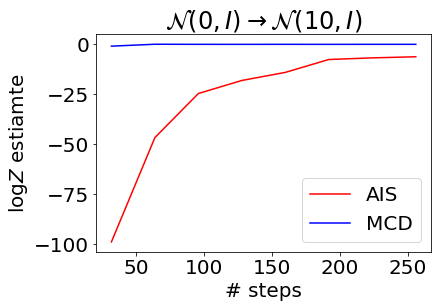}
  \end{center}
  \vspace{-0.5cm}
  \caption{Comparing $\log Z$ estimates as a function of $K$ using AIS and MCD. Both estimates use the same forward kernels but reweight samples in a different way using distinct backward kernels. Initial distribution $\pi_0$ is 20-dimensional $\mathcal{N}(0, I)$ and progressively shifts to the target $\pi=\pi_K=\mathcal{N}(10, I)$. The MCD estimate is much closer to the ground truth ($\log Z=0$) than AIS.}
  \label{fig:figure2_mcd_vs_ais}
  \vspace{-0.5cm}
\end{wrapfigure}

Another illustration of the suboptimality of AIS is to consider a scenario where the proposal is a homogeneous MCMC chain, i.e. $x_0 \sim \pi_0$ and $x_k \sim F(\cdot|x_{k-1})$ for $F$ a $\pi$-invariant MCMC kernel; i.e. use $F_k=F$ and $\pi_k=\pi$ for $k=1,...,K$. If $F$ is reasonably well-mixing, then $q_K \approx \pi$ for $K$ large enough and the evidence estimate $w_{\textup{mar}}(x_K)=\gamma(x_K)/q_K(x_K)$ should have small variance. However, it is easy to check that we have $w_{\textup{ais}}(x_{0:K})=\gamma(x_0)/\pi_0(x_0)$ for the exact same proposal; i.e. the AIS estimate does not depend on the MCMC samples $x_{1:K}$ and boils down to the IS estimate of $Z$ using the proposal $\pi_0$.

These two examples illustrate that it would be preferable to use $w_{\textup{mar}}(x_K)$ rather than $w_{\textup{ais}}(x_{0:K})$. In Appendix \ref{sev:varcalcul}, we provide a detailed comparison of both estimates in a scenario where their variance can be computed analytically. We propose in the next section an unbiased estimate of the evidence (MCD) approximating $w_{\textup{mar}}(x_K)$ based on a different choice of backward kernels. As illustrated in Figure \ref{fig:figure2_mcd_vs_ais}, significant gains can be achieved.

\section{Optimized Annealed Importance Sampling}
We show here that the optimal extended target distribution $P$ minimizing the variance of the evidence estimate (\ref{eq:evidenceestimate}) is defined through the time-reversal of the proposal $Q$. By exploiting a connection to SGM, we can approximate this reversal using score matching when the proposal is obtained through an unadjusted overdamped or underdamped Langevin algorithm.

\subsection{Optimal Extended Target Distribution via Time Reversal} We summarize here Proposition 1 of \cite{Del-Moral:2006}; see also \citep{sohl2015deep,bernton2019SBsamplers}. 

\begin{prop}\label{prop:optimalextentedtarget} For a proposal $Q$ of the form (\ref{eq:AISproposal}), the extended target $P$ of the form (\ref{eq:AIStarget}) minimizing both the Kullback--Leibler divergence $\KL(Q||P)$ and the variance of the evidence estimate $w(x_{0:K})=\Gamma(x_{0:K})/Q(x_{0:K})$ for $x_{0:K}\sim Q$ is given by $P^{\textup{opt}}(x_{0:K})=\Gamma^{\textup{opt}}(x_{0:K})/Z$ where
\begin{equation}
   \Gamma^{\textup{opt}}(x_{0:K})=\gamma(x_K)\prod_{k=0}^{K-1} B^{\textup{opt}}_k(x_k|x_{k+1}),\quad \quad B^{\textup{opt}}_k(x_k|x_{k+1})=\frac{q_{k}(x_{k})F_{k+1}(x_{k+1}|x_k)}{q_{k+1}(x_{k+1})}.
\end{equation}
In particular, one has 
\begin{equation}
  w_{\textup{mar}}(x_K)= \frac{\gamma(x_K)}{q_K(x_K)}=\frac{\Gamma^{\textup{opt}}(x_{0:K})}{Q(x_{0:K})},\quad\text{and}\quad \KL(Q||P^\textup{opt})=\KL(q_K||\pi).
\end{equation}
\end{prop}
This result follows simply from the chain rule and the law of total variance which yield
\begin{align}\label{eq:KLdecompomain}
    &\KL(Q||P)=\KL(q_K||\pi)+\E_{q_K}\Big[\KL(Q(\cdot|x_K)||P(\cdot|x_K))\Big],\\
    &\textup{var}_Q[w(x_{0:K})]=\textup{var}_{q_K}[w_{\textup{mar}}(x_{K})]+\E_{q_K}[\textup{var}_{Q(\cdot|x_K)}[w(x_{0:K})]].
\end{align}
Both quantities are clearly minimized by selecting $P(x_{0:K-1}|x_K)=Q(x_{0:K-1}|x_K)$. 

We emphasize that Proposition \ref{prop:optimalextentedtarget} applies to any forward kernels $(F_k)_{k=1}^{K}$ including MCMC kernels, unadjusted Langevin  kernels or even deterministic maps\footnote{The case of deterministic maps corresponds to normalizing flow components, where the inverse flow is the optimal and only valid reversal, see e.g. \citep{Arbel2021} which includes a detailed literature review.}. It shows that $P^{\textup{opt}}$ is the distribution of a backward process initialized at $\pi$ which then follows the time-reversed dynamics of the forward process $Q$. If we had $q_K=\pi$, then we would have $P^{\textup{opt}}=Q$ as then $P^{\textup{opt}}$ would correspond to the backward decomposition of $Q$.

\subsection{Time reversal, Score matching and ELBO for unadjusted overdamped Langevin} We concentrate here on the case where  $(F_k)_{k=1}^K$ correspond to a time-inhomogeneous unadjusted (overdamped) Langevin algorithm (ULA) as used in \citep{heng2017controlled,wunoe2020stochastic,thin2021MCVAE}; that is we consider $F_k(x_k|x_{k-1})=\mathcal{N}(x_k;x_{k-1}+\delta \nabla \log \pi_k(x_{k-1}),2\delta I)$ where $\delta>0$ is a stepsize. Let $\delta:=T/K$ then, as $K \rightarrow \infty$, the proposal $Q$ converges to the path measure $\mathcal{Q}$ of the following inhomogeneous Langevin diffusion $(x_t)_{t\in [0,T]}$ defined by the stochastic differential equation (SDE)
\begin{equation}\label{eq:nonhomogeneousLangevin}
    \rmd x_t=\nabla \log \pi_t(x_t) \rmd t+\sqrt{2} \rmd B_t,\quad\quad x_0\sim \pi_0,
\end{equation}
where $(B_t)_{t\in [0,T]}$ is standard multivariate Brownian motion and we are slightly abusing notation from now on as $\pi_t$ for $t=t_k=k\delta$ corresponds to $\pi_k$ in discrete-time. Many quantitative results measuring the discrepancy between the law of $x_T$ and $\pi_T=\pi$ for such annealed diffusions have been obtained; see e.g. \citep{fournier2021simulated,tang2021simulated}. From \citep{haussmann1986time} , it is known that the time-reversed process $(\bar{x}_t)=(x_{T-t})_{t\in[0,T]}$ is also a diffusion given by 
\begin{equation}\label{eq:Langevinreversal}
    \rmd \bar{x}_t=\big\{-\nabla \log \pi_{T-t}(\bar{x}_t)+2 \nabla \log q_{T-t}(\bar{x}_t)\big\} \rmd t+\sqrt{2} \rmd \bar{B}_t,\quad\quad \bar{x}_0\sim q_T,
\end{equation}
where $(\bar{B}_t)_{t\in [0,T]}$ is another multivariate Brownian motion.
The continuous-time version of $P^{\textup{opt}}$ is the path measure $\mathcal{P}^{\textup{opt}}$ defined by the diffusion (\ref{eq:Langevinreversal}) but initialized at $\bar{x}_0 \sim \pi$ rather than $q_T$ as noted in \cite{bernton2019SBsamplers}; see Figure \ref{fig:Langevin} for an illustration. This shows that approximating $(B^{\textup{opt}}_k)_{k=0}^{K-1}$ requires approximating the so-called scores $(\nabla \log q_t(x))_{t\in [0,T]}$. This can be derived heuristically through the fact that a Taylor expansion yields the following approximation of the optimal backward kernels, $B^{\textup{opt}}_k(x_k|x_{k+1})\approx \mathcal{N}(x_k;x_{k+1}-\delta \nabla \log \pi_{k+1}(x_{k+1})+2 \delta \nabla \log q_{k+1}(x_{k+1}),2\delta I)$, which indeed corresponds to a Euler discretization of (\ref{eq:Langevinreversal}); see e.g. \cite[Section 2.1]{debortoli2021diffusion}.

In SGM \citep{song2020score}, one gradually adds noise to data using an Ornstein--Ulhenbeck diffusion to transform the complex data distribution into a Gaussian distribution and the generative model is obtained by approximating the time-reversal of this diffusion initialized by Gaussian noise. Practically, the time-reversal approximation is obtained by estimating the scores of the noising diffusion using denoising score matching \citep{vincent2011connection}. While in our setup, the diffusion (\ref{eq:nonhomogeneousLangevin}) instead goes from a simple distribution to a complex one (see Appendix \ref{sec:sgmlangevinaisdiffusions} for a discussion), we can still use score matching ideas. We define a path measure $\mathcal{P}_\theta$ approximating $\mathcal{P}^{\textup{opt}}$ using a neural network $ s_\theta(T-t,\bar{x}_t)$ in place of $\nabla \log q_{T-t}(\bar{x}_t)$ in (\ref{eq:Langevinreversal}), i.e. we consider 
\begin{equation}\label{eq:Langevinreversalapprox}
  \rmd \bar{x}_t=\big\{-\nabla \log \pi_{T-t}(\bar{x}_t)+2 s_\theta(T-t,\bar{x}_t)\big\} \rmd t+\sqrt{2} \rmd \bar{B}_t,\quad\quad \bar{x}_0\sim \pi.
\end{equation}
We would like to learn $\theta$ by minimizing $\KL(\mathcal{Q}||\mathcal{P}_\theta)$ over $\theta$, i.e.\ equivalently we maximize a continuous-time ELBO. Note that it is neither easily feasible to minimize $\KL(\mathcal{P}^{\textup{opt}}||\mathcal{P}_\theta)$ (as one cannot sample from $\pi$) nor it is desirable as the evidence estimate is computed using samples from $\mathcal{Q}$. Hence we want the scores to be well-approximated in regions of high-probability mass under $\mathcal{Q}$.%

In practice, the diffusions corresponding to $\mathcal{Q}$ and $\mathcal{P_\theta}$  have to be discretized, so a more direct route adopted here is to simply take inspiration from (\ref{eq:Langevinreversal}) and to consider the parameterized backward kernels $B^{\theta}_k(x_k|x_{k+1})=\mathcal{N}(x_k;x_{k+1}-\delta \nabla \log \pi_{k+1}(x_{k+1})+2 \delta s_\theta(k+1,x_{k+1}),2\delta I)$ to obtain a parameterized extended target $P_\theta$ and corresponding unnormalized target $\Gamma_\theta$. We then learn $\theta$ by minimizing $\KL(Q||P_\theta)$ where 
\begin{equation*}
 \textstyle 
 Q(x_{0:K})=\pi_0(x_0) \prod_{k=0}^{K-1} F_{k+1}(x_{k+1}|x_k),\quad
  P_\theta(x_{0:K})=\pi(x_K)\prod_{k=0}^{K-1}B^{\theta}_{k}(x_{k}|x_{k+1}).
 \end{equation*}
This is obviously equivalent to maximizing the ELBO $\mathbb{E}_Q[\log w_\theta(x_{0:K})]$ where $w_\theta(x_{0:K})=\Gamma_\theta(x_{0:K})/Q(x_{0:K})$. We note that it has previously been proposed to learn parameterized backward kernels for general AIS proposals \citep{salimans2015markov,huang2018improving}. However, the parameterization adopted therein, $B^{\theta}_k(x_k|x_{k+1})=\mathcal{N}(x_k;\mu_\theta(x_{k+1}),\Sigma_\theta(x_{k+1}))$, does not leverage the structure of the true reversal and performs poorly experimentally \cite[Section 4.2]{thin2021MCVAE}.

As established in the next proposition, the continuous and discrete time approaches coincide for  $\delta \ll 1$. Once $\theta$ is learned, we then obtain an unbiased estimate of $Z$ through $w_\theta(x_{0:K})$ for $x_{0:K}\sim Q$.

\begin{prop}\label{prop:ELBOmixed}Under regularity conditions, we have 
\begin{align}
 \KL(\mathcal{Q}||\mathcal{P}_\theta)&= \E_{\mathcal{Q}}\Big[\int_0^T||s_{\theta}(t,x_t)-\nabla \log q_t(x_t)||^2\rmd t\Big]+C_1 \label{eq:scorematchingpropa}\\
    &=\sum_{k=1}^{K}\int_{t_{k-1}}^{t_{k}}\E_\mathcal{Q}\left[||s_{\theta}(t,x_{t})-\nabla \log q_{t|t_{k-1}}(x_{t}|x_{t_{k-1}})||^{2}\right]\rmd t+C_2, \label{eq:scorematchingpropb}
\end{align}
where $t_k=k \delta$, $K=T/\delta$, $q_{t|s}(x'|x)$ is the density of $x_{t}=x'$ given $x_s=x$ under $\mathcal{Q}$ and $C_1,C_2$ constants independent of $\theta$. Let $\mathcal{L}(\theta)=\delta \sum_{k=1}^{K} \E_Q\left[||s_{\theta}(k,x_k)-\nabla \log F_k(x_{k}|x_{k-1})||^{2}\right]$ denote a discrete-time approximation of this loss. We have $ \nabla \KL(Q||P_\theta)=\nabla \mathcal{L}(\theta)+\epsilon(\theta)$ for some function $\epsilon$ satisfying $\lim_{K\rightarrow \infty}\epsilon(\theta)=0$.

\end{prop}

Equation (\ref{eq:scorematchingpropa}) shows that $\KL(\mathcal{Q}||\mathcal{P}_\theta)$ corresponds to a score matching loss as for SGM \citep{song2021maximum}. It is possible to rewrite this loss as (\ref{eq:scorematchingpropb}) so as to replace the intractable score term $\nabla \log q_t(x_t)$ by the easy to approximate gradients of the log-transitions $\nabla \log q_t(x_t|x_{t_{k-1}})$ \citep{vincent2011connection}. In practice, as mentioned above, we simply learn $\theta$ by minimizing the discrete-time KL discrepancy $\KL(Q||P_\theta)$. This formulation is also very convenient as we can additionally learn potential parameters $\phi$ of a $Q_{\phi}$ using the same criterion. Note from equation (\ref{eq:KLdecompomain}) that the KL divergence decomposes as firstly a term penalizing the difference in the approximating measure and the fixed target at the final time and secondly another term which can be reduced by optimization of both $Q_{\phi}$ and $P_{\theta}$ conditioned on $x_{T}$; see e.g. \citep{Agakov2004}.

Pseudo-code for our approach (with a comparison to the AIS algorithm proposed in \cite{heng2017controlled,wunoe2020stochastic,thin2021MCVAE}) can be found in Algorithm \ref{alg:ulaais}.

\setlength{\textfloatsep}{0.1cm}
\begin{algorithm}
\caption{Unadjusted Langevin {\textcolor{red}{AIS}}/{\textcolor{blue}{MCD}} --  \textcolor{red}{red} instructions for AIS and \textcolor{blue}{blue} for MCD}\label{alg:ulaais}
\begin{algorithmic}
\Require Unnormalized target $\gamma(x)$, initial state proposal $\pi_0(x)$, number steps $K$, stepsize $\delta$, annealing schedule $\{\beta_k\}_{k=0}^K$, {\textcolor{blue}{score model $s_\theta(k, x)$}}
\State Sample $x_0 \sim \pi_0(x_0)$
\State Set $\log w = -\log \pi_0(x_0)$
\For{$k = 1$ to $K$}
\State Define $\log \gamma_k(\cdot) = \beta_k \log \gamma(\cdot) + (1 - \beta_k) \log \pi_0(\cdot)$
\State Define $F_k(x_k | x_{k-1}) = \mathcal{N}(x_k; x_{k-1} + \delta \nabla \log \gamma_k(x_{k-1}), 2 \delta I)$
\State Sample $x_k \sim F_k(\cdot| x_{k-1})$

\State \textcolor{red}{Define $B_{k-1}(x_{k-1} | x_k) = F_k(x_{k-1} | x_{k})$ } \Comment{\textcolor{red}{AIS}}
\State {\textcolor{blue}{Define $B_{k-1}(x_{k-1} | x_k) = \mathcal{N}(x_{k-1}; x_k - \delta \nabla \log  \gamma_k(x_k) + 2 \delta s_\theta(k, x_k), 2 \delta I)$ }}\Comment{{\textcolor{blue}{MCD}}}

\State Set $\log w = \log w + \log B_{k-1}(x_{k-1} | x_k) - \log F_k(x_{k} | x_{k-1})$
\EndFor
\State Set $\log w = \log w + \log \gamma(x_K)$
\end{algorithmic}
\end{algorithm}
\setlength{\floatsep}{0.1cm}

\subsection{Incorporating Hamiltonian dynamics via the underdamped Langevin equation}\label{section:HamiltonianMain}
We now consider a proposal defined on an extended space which arises from the time-discretization of a time-inhomogeneous underdamped Langevin dynamics; see e.g. \cite[Chapter 6]{leimkuhler2016molecular}. In this scenario, we first focus on continuous-time as the development of suitable numerical integrators is much more involved than for overdamped Langevin diffusions. We consider the diffusion  $(x_t,p_t)_{t\in [0,T]}$ where $p_t \in \mathbb{R}^d$ is a momentum variable
\begin{equation}\label{eq:underdampedLangevin}
\rmd x_{t} =M^{-1}p_{t}\rmd t,\qquad
\rmd p_{t}=\nabla \log \pi_{t}(x_{t})\rmd t-\zeta p_{t}\rmd t+\sqrt{2\zeta}M^{1/2}\rmd B_{t}
\end{equation}
initialized at $x_0\sim \pi_0, p_0 \sim \mathcal{N}(0,M)$ defining the path measure $\mathcal{Q}$. Here $M$ is a positive definite mass matrix, $\zeta>0$ a friction coefficient and $(B_t)_{t\in [0,T]}$ a multivariate Brownian motion. If $\pi_t$ was not time-varying, e.g. $\pi_t=\pi$, the invariant distribution of this diffusion would be given by $\bar{\pi}(x,p)=\pi(x)\mathcal{N}(p;0,M)$. Intuitively, in the time varying case the SDE will have enough time to approximate each intermediate $\bar{\pi}_t(x,p)=\pi_t(x)\mathcal{N}(p;0,M)$ if we change the target sufficiently slowly.  We can think of underdamped Langevin as a continuous-time version of Hamiltonian dynamics with continuous stochastic partial momentum refreshment \citep{horowitz1991generalized}.

From \citep{haussmann1986time}, the time-reversal of the diffusion (\ref{eq:underdampedLangevin}) is also a diffusion process $(\bar{x}_t,\bar{p}_t)_{t\in [0,T]}=(x_{T-t},p_{T-t})_{t\in[0,T]}$  given by $(\bar{x}_0,\bar{p}_0)\sim \eta_T$ and
\begin{align}\label{eq:timereversalnonhomogeneousunderdampedLangevinpositivetime}
\rmd \bar{x}_{t} & =-M^{-1}\bar{p}_{t}\rmd t,\\
\rmd \bar{p}_{t} & =-\nabla \log \pi_{T-t}(\bar{x}_{t})\rmd t+\zeta \bar{p}_{t}\rmd t+2\zeta M\nabla_{\bar{p}_{t}}\log \eta_{T-t}(\bar{x}_{t},\bar{p}_{t})\rmd t+\sqrt{2\zeta}M^{1/2}\rmd \bar{B}_{t},\nonumber
\end{align}
where $\eta_t$ denotes the density $(x_t,p_t)$ under (\ref{eq:underdampedLangevin}). In this case, the continuous-time version of $P_{\textup{opt}}$ is the path measure $\mathcal{P}_{\textup{opt}}$ defined by the diffusion (\ref{eq:timereversalnonhomogeneousunderdampedLangevinpositivetime}) but initialized at $\bar{x}_0 \sim \pi, \bar{p}_0 \sim \mathcal{N}(0,M)$ rather than $\eta_T$. We will approximate it by the path measure $\mathcal{P}_\theta$ using a neural network $ s_\theta(T-t,\bar{x}_t,\bar{p}_t)$ in place of $\nabla \log \eta_{T-t}(\bar{x}_t,\bar{p}_t)$ in (\ref{eq:timereversalnonhomogeneousunderdampedLangevinpositivetime}), i.e. we consider
\begin{align}\label{eq:approximatetimereversalnonhomogeneousunderdampedLangevinpositivetime}
\rmd \bar{x}_{t} & =-M^{-1}\bar{p}_{t}\rmd t,\\
\rmd \bar{p}_{t} & =-\nabla \log \pi_{T-t}(\bar{x}_{t})\rmd t+\zeta \bar{p}_{t}\rmd t+2\zeta M s_\theta(T-t,\bar{x}_t,\bar{p}_t)\rmd t+\sqrt{2\zeta}M^{1/2}\rmd \bar{B}_{t}.\nonumber
\end{align}

As for overdamped Langevin, we could also learn $\theta$ by minimizing $\KL(\mathcal{Q}||\mathcal{P}_\theta)$ over $\theta$. This again corresponds to minimizing a score matching loss albeit of a form slightly different from  (\ref{eq:scorematchingpropa}).
\begin{prop}\label{prop:ELBOunderdampedCT}Under regularity conditions, we have 
\begin{align}
 \textstyle   \KL(\mathcal{Q}||\mathcal{P}_\theta)&= \zeta \E_{\mathcal{Q}}\Big[\int_0^T||s_{\theta}(t,x_t,p_t)-\nabla_{p_t} \log \eta_t(x_t,p_t)||^2\rmd t\Big]+C_1\label{eq:scorematchingunderdamped1}\\
    &=\zeta \sum_{k=1}^{K}\int_{t_{k-1}}^{t_{k}}\E_\mathcal{Q}\left[||s_{\theta}(t,x_{t},p_t)-\nabla_{p_t} \log \eta_{t|t_{k-1}}(x_{t},p_t|x_{t_{k-1}},p_{t_{k-1}})||_M^{2}\right]\rmd t+C_2, \nonumber
\end{align}

where $||x||_M:=u^{\textup{T}}M u$, $t_k=k \delta$, $K=T/\delta$, $C_1,C_2$ are constants independent of $\theta$ and $\eta_{t|s}(x',p'|x,p)$ is the density of $(x_{t},p_t)=(x',p')$ given $(x_s,p_s)=(x,p)$ under $\mathcal{Q}$. 
\end{prop}

\begin{algorithm}[t]
\caption{Unadjusted Hamiltonian \textcolor{red}{AIS}/\textcolor{blue}{MCD}  --  \textcolor{red}{red} instructions for AIS and \textcolor{blue}{blue} for MCD}\label{alg:uhaais}
\begin{algorithmic}
\Require Unnormalized target $\gamma(x)$, initial state proposal $\pi_0(x)$, number steps $K$, stepsize $\eta$, annealing schedule $\{\beta_k\}_{k=0}^K$, damping coefficient $h$, mass matrix $M$, \textcolor{blue}{score model $s_\theta(k, x, p)$}
\State Sample $x_0 \sim \pi_0(x_0)$ and $p_0 \sim \mathcal{N}(p_0; 0, M)$
\State Set $\log w = -\log \pi_0(x_0) - \log \mathcal{N}(p_0; 0, M)$
\For{$k = 1$ to $K$}
\State Define $\log \gamma_k(\cdot) = \beta_k \log \gamma(\cdot) + (1 - \beta_k) \log \pi_0(\cdot)$
\State Sample $\tilde{p}_{k} \sim \mathcal{N}(h p_{k-1}, (1-h^2)M)$
\State Set $\mu_q = p_{k-1}$
\State \textcolor{red}{Set $\mu_p = \tilde{p}_k$} \Comment{\textcolor{red}{UHA reversal mean}}
\State \textcolor{blue}{Set $\mu_p = \tilde{p}_k - 2 \log (h) [M s_\theta(k, x_{k-1}, \tilde{p}_k) + \tilde{p}_k]$} \Comment{\textcolor{blue}{MCD reversal mean}}

\State Set $\log w = \log w + \log \mathcal{N}(p_{k-1}; h\mu_p, (1-h)^2 M) - \log \mathcal{N}(\tilde{p}_{k}; h\mu_q, (1-h)^2 M)$

\State Run leapfrog integrator on $\gamma_k$ and set $(x_k, p_k) = \Phi(x_{k-1}, \tilde{p}_{k})$
\EndFor
\State Set $\log w = \log w + \log \gamma(x_K) + \log \mathcal{N}(p_K; 0, M)$
\end{algorithmic}
\end{algorithm}

While the continuous-time perspective shed light on how to parameterize an approximation to the time-reversal, this does not lead directly to an implementable discrete-time algorithm for underdamped Langevin. Contrary to overdamped Langevin, we cannot indeed simply use an Euler discretization of (\ref{eq:underdampedLangevin}) defining $\mathcal{Q}$ and (\ref{eq:approximatetimereversalnonhomogeneousunderdampedLangevinpositivetime}) defining $\mathcal{P}_\theta$ to obtain some discrete-time forward and backward kernels and then compute $w_\theta(x_{0:K},p_{0:K})=\Gamma_\theta(x_{0:K},p_{0:K})/Q(x_{0:K},p_{0:K})$. This is because this ratio is not well-defined due to the lack of noise on the position component in both (\ref{eq:underdampedLangevin}) and (\ref{eq:approximatetimereversalnonhomogeneousunderdampedLangevinpositivetime}). 

The integrator we use for the forward equation (\ref{eq:underdampedLangevin}), consists in alternating partial momentum refreshments and deterministic leapfrog steps (see e.g \citep{leimkuhler2016molecular, neal2011mcmc}) giving
\begin{align}
\tilde{p}_{t+\delta} \sim \mathcal{N}(h p_t, (1-h^2)M),\qquad
(x_{t+\delta},p_{t+\delta})=\Phi_t(x_t,\tilde{p}_{t+\delta}),    
\end{align}
with $h=\exp\lbrace -\zeta \delta \rbrace$ and $\Phi_t$ is the leapfrog integrator for $\pi_t$. The resulting forward sampler is similar to the one proposed by \cite{Geffner:2021}, except we do not flip the momentum after the leapfrog step\footnote{\cite{Geffner:2021} were not attempting to discretize an underdamped Langevin dynamics.}. This integrator may be interpreted as a splitting method for equation (\ref{eq:underdampedLangevin}); see e.g \cite[Chapter 7]{leimkuhler2016molecular}. 

We need the integrator for the reversal to fulfill two criteria. First, by definition, as the time step $\delta \to 0$ it must recover the SDE (\ref{eq:approximatetimereversalnonhomogeneousunderdampedLangevinpositivetime}). Second, the importance weight of the forward sampler to the reversal must be well defined. Since the leapfrog integrator is a diffeomorphism (or flow) the only possible way to get a well defined reversal for these steps is to take the inverse $\Phi_t^{-1}$. As the transformation is also volume preserving, the contribution from the deterministic forward and reverse terms will then exactly cancel in the importance weight. The required form of the reverse integrator is
\begin{align}
(x_t,\tilde{p}_{t+\delta})=\Phi_{t}^{-1}(x_{t+\delta},p_{t+\delta}),\qquad p_t  &\sim \mathcal{N}(h f_{\theta}(t+\delta,x_t,\tilde{p}_{t+\delta}), (1-h^2)M),
\end{align}
where $f_{\theta}(t+\delta,x_t,\tilde{p}_{t+\delta}):= \widetilde{p}_{t+\delta}+\delta2\zeta[Ms_{\theta}(t,x_{t},\widetilde{p}_{t+\delta})+\widetilde{p}_{t+\delta}]$. In Appendix \ref{app:reverseintegratorHamiltonian} we show that as $\delta \to 0$ this can indeed be interpreted as a valid split integrator for the reverse SDE (\ref{eq:approximatetimereversalnonhomogeneousunderdampedLangevinpositivetime}). The crucial point of algorithmic difference from \citep{dai2020invitation,Geffner:2021,Zhang2021} arises from our necessary form for the mean of the reverse momentum refreshment. These works use $h \tilde{p}_{t+\delta}$ as the mean instead of $h f_{\theta}(t+\delta,x_t,\tilde{p}_{t+\delta})$. 
We again transition to discretized notation with $\delta:=T/K$, and $k=0,...,K$. In this case, the log importance weight, corresponding to the log evidence estimate, satisfies
\begin{equation}\label{eq:AISOPTestimateHamiltonian}
\log w_\theta(\textbf{x},\textbf{p})=\log \frac{\gamma(x_K)\mathcal{N}(p_K;0,M)}{ \pi_0(x_0)\mathcal{N}(p_0;0,M)}+\sum_{k=1}^{K} \log \frac{\mathcal{N}(p_{k-1};h f_{\theta}(k,x_{k-1},\tilde{p}_{k}),(1-h^2)M)}{\mathcal{N}(\tilde{p}_{k};h p_{k-1},(1-h^2)M)},
\end{equation}

where $(\textbf{x},\textbf{p})$ denote all the variables introduced by our integration scheme. We can show informally that minimizing $\KL(Q||P_\theta)$, i.e. maximizing the ELBO given $\mathbb{E}_Q[\log w_\theta(\textbf{x},\textbf{p})]$, again corresponds to minimizing a score matching type loss (\ref{eq:scorematchingunderdamped1}) when $\delta \ll 1$; see Appendix \ref{app:reverseintegratorHamiltonian}. Pseudo-code for our approach can be found in Algorithm \ref{alg:uhaais}.

\section{Experiments}
We run a number of experiments estimating normalizing constants to validate our approach, MCD and compare to differentiable AIS with ULA~\citep{wunoe2020stochastic,thin2021MCVAE} and Unadjusted Hamiltonian Annealing (UHA)~\citep{Geffner:2021, Zhang2021}. We first investigate the performance of these approaches on static target distributions using the same, fixed initial distribution and annealing schedule.
Finally, we explore the performance of the methods for VAEs. Here, being the most expensive of our experiments, we include runtime comparisons of our method compared to baselines. Additional results on a Normalizing Flow target can be found in Appendix \ref{app:flowtarg}. Full experimental details, chosen hyper-parameters, and model architectures can be found in Appendix \ref{sec:expe}.

Our score model is parameterized by an MLP with residual connections that is conditioned on integration time $t$, and on the momentum term for the Hamiltonian case (see Algorithm \ref{alg:uhaais}). For an ablation on various network architectures we refer the reader to Appendix \ref{app:network_ablation}.

\subsection{Static Targets} We estimate the normalizing constants of five simple distributions with known normalizing constants equal to $\log Z=0$; $\mathcal{N}(10, I)$, $\mathcal{N}(0, 0.1 I )$, a Gaussian mixture with 8 components whose means are drawn from $\mathcal{N}(3, I)$ where each component has variance 1, a standard Laplace distribution and a Student's T distribution with 3 degrees of freedom. We use $\mathcal{N}(0, I)$ as our initial distribution for all targets except for the Gaussian mixture and $\mathcal{N}(0, 0.1 I )$ which both use $\mathcal{N}(0, 3^2 I )$. We run each method using $K \in \{64,256\}$ steps and use a fixed, linear annealing schedule. For all methods, sampling step-sizes per-timestep are tuned to via gradient descent to maximize the ELBO and the diagonal mass matrix is learned for the Hamiltonian samplers. Gaussian Mixture and Student-T results can be found in Tables \ref{tab:staticmog} and \ref{tab:staticstudentt}, respectively. Additional results can be found in Appendix \ref{app:static}.

\begin{table}[h]
    \centering
    \begin{tabular}{c ||c|c||c|c || c  | c || c  | c }
    \hline
     Sampler  & \multicolumn{2}{c}{ULA}&  \multicolumn{2}{c}{UHA}& \multicolumn{2}{c}{ULA-MCD} & \multicolumn{2}{c}{UHA-MCD} \\
     $\#$ steps  &64&256  &64&256 &64&256  &64&256 \\
    \hline
    \hline
    Dim-20 & \makecell{-0.47 \\ $\pm$ 0.22} &  \makecell{-0.13 \\ $\pm$ 0.07} & \makecell{-0.03 \\ $\pm$ 0.18} &  \makecell{0.015 \\ $\pm$ 0.03} & \makecell{0.01 \\ $\pm$ 0.02} &  \makecell{0.01 \\ $\pm$ 0.01} & \makecell{0.01 \\ $\pm$ 0.02} & \makecell{-0.01 \\ $\pm$ 0.01} \\
\hline
Dim-200 & \makecell{-85.62 \\ $\pm$ 2.01} &  \makecell{-21.98 \\ $\pm$ 1.35} & \makecell{-8.20 \\ $\pm$ 1.84} &  \makecell{-1.26 \\ $\pm$ 0.35} & \makecell{-0.25 \\ $\pm$ 0.03} &  \makecell{0.08 \\ $\pm$ 0.122} & \makecell{0.20 \\ $\pm$ 0.49} &  \makecell{-0.05 \\ $\pm$ 0.04} \\
\hline
Dim-500 & \makecell{-304.06 \\ $\pm$ 5.48} &  \makecell{-83.50 \\ $\pm$ 7.88} & \makecell{-44.45 \\ $\pm$ 3.24} &  \makecell{-8.60 \\ $\pm$ 1.69} & \makecell{-2.61 \\ $\pm$ 1.26} &  \makecell{1.01 \\ $\pm$ 0.99} & \makecell{-1.74 \\ $\pm$ 1.25} &  \makecell{-1.02 \\ $\pm$ 0.03} \\
    \hline
    \end{tabular}
    \caption{$\log Z$ estimates for a Gaussian mixture target. Averages and standard errors over 3 seeds.}
    \label{tab:staticmog}
\end{table}

On average and as expected, UHA outperforms ULA on most targets. Further our approximation to the optimal backward kernels can yield considerable improvements -- enabling the ULA sampler to produce better results than UHA (using the standard AIS backward kernels). Additionally, we see that our approximation to the optimal backward kernels improves the performance of UHA as well.

We further emphasize that often ULA-MCD and UHA-MCD with 64 steps outperform or is on-par with ULA and UHA with 256 steps, a difference of factor 4 in terms of target gradient evaluations.
As we show below in \Cref{tab:vae}, the additional computational costs of fitting the score model of our method are only roughly twice that of the baselines.

\begin{table}[h]
    \centering
    \begin{tabular}{c ||c|c||c|c || c  | c || c  | c }
    \hline
     Sampler  & \multicolumn{2}{c}{ULA}&  \multicolumn{2}{c}{UHA}& \multicolumn{2}{c}{ULA-MCD} & \multicolumn{2}{c}{UHA-MCD} \\
     $\#$ steps  &64&256  &64&256 &64&256  &64&256 \\
    \hline
    \hline
    Dim-20 & \makecell{-0.09 \\ $\pm$ 0.02} & \makecell{-0.02 \\ $\pm$ 0.01} & \makecell{-0.04 \\ $\pm$ 0.02} & \makecell{-0.02 \\ $\pm$ 0.01} & \makecell{-0.06 \\ $\pm$ 0.02} & \makecell{-0.00 \\ $\pm$ 0.01} & \makecell{-0.03 \\ $\pm$ 0.04} & \makecell{-0.01 \\ $\pm$ 0.01} \\
\hline
Dim-200 & \makecell{-1.63 \\ $\pm$ 0.23} & \makecell{-0.88 \\ $\pm$ 0.15} & \makecell{-0.36 \\ $\pm$ 0.34} & \makecell{-0.10 \\ $\pm$ 0.10} & \makecell{-0.82 \\ $\pm$ 0.28} & \makecell{-0.30 \\ $\pm$ 0.38} & \makecell{-0.48 \\ $\pm$ 0.20} & \makecell{-0.10 \\ $\pm$ 0.07} \\
\hline
Dim-500 & \makecell{-5.43 \\ $\pm$ 0.78} & \makecell{-3.10 \\ $\pm$ 0.07} & \makecell{-2.86 \\ $\pm$ 0.36} & \makecell{-1.03 \\ $\pm$ 0.13} & \makecell{-4.00 \\ $\pm$ 0.51} & \makecell{-2.23 \\ $\pm$ 0.18} & \makecell{-2.03 \\ $\pm$ 0.18} & \makecell{0.06 \\ $\pm$ 0.30} \\
\hline
    \hline
    \end{tabular}
    \caption{$\log Z$ estimates for a Student-T target. Averages and standard errors over 3 seeds.}
    \label{tab:staticstudentt}
\end{table}

\subsection{Application to Amortized Inference} Next, we explore the application of our method to amortized inference, in the context of VAEs \citep{kingma2013auto, rezende+al:2014:icml}.
These models are trained to infer latent representations using an inference neural network that consumes an input and produces parameters of an approximation to the true posterior distribution of the underlying generative model.
In particular, this posterior distribution is different for each input.
When applying AIS to VAE inference \citep{thin2021MCVAE, Geffner:2021, Zhang2021}, the output of the inference network parameterizes the initial distribution of the annealing sequence to the true posterior.
By training this end-to-end, we effectively learn the  initial distribution for the diffusion process.
Consequently, the diffusion marginals and their score vectors $\nabla \log q_t$ are different for every input, and we need to condition our score model $s_\theta$ on the inputs to reflect that.
We achieve this by projecting the last hidden layer of the inference network into a summary vector that is concatenated to the other conditioning inputs of $s_\theta$.

We train a  VAE on the binarized MNIST dataset \citep{salakhutdinov2008quantitative}, re-using architectures proposed in \citep{Geffner:2021, Burda2015} (two-layer MLP encoder/decoder, Bernoulli likelihood). All generative models use the same architecture and hyper-parameters. We compare standard amortized variational inference with annealed ULA and UHA with standard AIS backward transition kernels, as well as ULA and UHA with our MCD transition kernels.
We match the number of sampler steps between ULA/ULA-MCD and UHA/UHA-MCD to 64 and 32 respectively.
ELBO and log-likelihood values on the test set are presented in Table \ref{tab:vae}.

\begin{table}[h]
    \centering
    \begin{tabular}{c |c | c | c | c| c }
    \hline
     Sampler  & VI & \multicolumn{1}{c}{ULA}&   \multicolumn{1}{c}{UHA} & \multicolumn{1}{c}{ULA-MCD}& \multicolumn{1}{c}{UHA-MCD} \\
    \hline
    \hline
ELBO & \makecell{-96.32 $\pm$ 0.40} & \makecell{-90.41 $\pm$ 0.17} & \makecell{-88.58 $\pm$ 0.51} & \makecell{-90.10 $\pm$ 0.10} & \makecell{{-88.08 $\pm$ 0.07}} \\
Log-lik.  & \makecell{-69.35 $\pm$ 0.36} & \makecell{-62.43 $\pm$ 0.25} & \makecell{-61.05 $\pm$ 1.84} & \makecell{-61.83 $\pm$ 0.42} & \makecell{{-58.58 $\pm$ 0.34}} \\
\hline
Iteration time & \makecell{0.024s} & \makecell{0.055s} & \makecell{0.050s} & \makecell{0.101s} & \makecell{0.098s} \\

Total time & \makecell{3263.40s} & \makecell{4694.52s} & \makecell{5072.19s} & \makecell{11304.36s} & \makecell{11345.59s} \\

\hline
    \hline
    \end{tabular}
    \caption{Test set performance for MNIST VAE. Averages and standard errors over 5 seeds. We additionally report runtimes for a single training iteration, and total experiment time.}
    \label{tab:vae}
\end{table}
\vspace{-.3cm}

We see that, as reported in prior works~\citep{Geffner:2021,thin2021MCVAE,Zhang2021}, Monte Carlo based inference methods provide a significant benefit over standard amortized variational inference.
In addition, our learned backward kernels lead to improved performance of both ULA and UHA, with the difference being more distinct for the latter.
We also note that UHA has significantly larger error bars than all other methods, but UHA-MCD does not appear to inherit this.
We further provide a runtime comparison in \Cref{tab:vae}, both for a single training iteration, and total experiment time (including evaluation).
We find that fitting the score models of our method results in roughly twice as much runtime (recall we fixed the number of sampler steps).

\section{Limitations}\label{section:limitations}
While we have demonstrated that our optimized backward kernels can lead to large improvements over standard AIS backward kernels, our proposed approach has a number of limitations. 
First, since our method relies upon unadjusted Langevin or Hamiltonian sampling, we inherit many of the issues with these approaches. Using unadjusted samplers enables a fully-differentiable ELBO estimate which can in theory be used to tune the many parameters of the sampler. However, these samplers require repeated gradient steps as an inner-loop which can lead to divergence dynamics and numerical instability, making optimization difficult. Next, our proposed training procedure has a notable increase in memory consumption as we store the entire sampling trajectory in memory to train our neural network. We further note that it is possible to compute our importance weights online using $\mathcal{O}(1)$ memory. We could utilize this with Monte Carlo sub-sampling of the timestep (as with SGM~\citep{ho2020denoising}) to derive a constant memory training procedure for our approximate time-reversal at the cost of increased variance but we leave exploring this to future work. Finally, we observed that, without mitigation, parameterized forward samplers $Q_{\phi}$ sometimes dropped modes during training. Since this is a known challenge of using reverse KL, this might be improved by using alternative optimization objectives \citep{Masrani2019,Matthews2022}.

\section{Discussion}
In this work we have explored AIS using unadjusted Langevin and Hamiltonian dynamics. We have demonstrated that the backward transition kernels typically used are suboptimal and we have presented the form of the optimal variance-minimizing backward kernels. We have further shown how an approximation to these kernels can be learned using score matching and that this objective corresponds to maximizing the ELBO in the limit of infinitesimally small time-discretization step-sizes. We have illustrated the benefit of using our proposed optimized backward kernels on a number of inference problems including fixed targets and amortized tasks with model learning. 

\newpage
\bibliography{references}
\bibliographystyle{apalike}

\newpage
\appendix
\section{Variance calculations}\label{sev:varcalcul}
We provide here, for a simple example, the variance expressions for $\log w_{\textup{ais}}(x_{0:K})$ and $\log w_{\textup{mar}}(x_{K})$. All the expectations in this section are w.r.t. the proposal $Q$.

We consider the scenario where $\pi_{0}(x)=\gamma_{0}(x)=\mathcal{N}(x;0,\sigma_0^2)$ and for $k\geq 1$
\[
\pi_{k}(x)=\mathcal{N}(x;0,\sigma_{k}^{2}),\quad \gamma_{k}(x)=\exp\left(-\frac{x^{2}}{2\sigma_{k}^{2}}\right). 
\]
We select the sequence of variances as follows
\[
\sigma_{k}^{2}=\left(\sigma_{0}^{2}\right)^{1-\frac{k}{K}}\left(\sigma^{2}\right)^{\frac{k}{K}}=\left(\frac{\sigma^{2}}{\sigma_{0}^{2}}\right)^{1/K}\sigma_{k-1}^{2}:=\beta_{K}\sigma_{k-1}^{2}
\]
so that $\pi(x)=\pi_{K}(x)=\mathcal{N}(x;0,\sigma^{2})$. We will pick $\sigma^{2}<\sigma_{0}^{2}$ so $\beta_{K}<1$.
Finally we consider the following proposal $Q$. At initialization $x_{0}\sim \pi_0$ and for $k\geq 1$ we have the following transitions kernels 
\[
F_{k}(x'|x)=\mathcal{N}(x';\alpha x,(1-\alpha^{2})\sigma_{k}^{2})
\]
 which are $\pi_{k}$-invariant and $\alpha$ determines how fast
it mixes. For $\alpha=0$, we have exact samples from $\pi_{k}$ and
as $\alpha\rightarrow1$ we are mixing less and less.
For this setup, it is possible to provide exact calculations for the expectation and variance of the log-evidence estimate.
\begin{prop}\label{prop:varcal} 
Under $Q$, we have $x_{k}\sim\mathcal{N}(0,\xi_{k}^{2})$
where $\xi_{0}^{2}=\sigma_{0}^{2}$ and
\begin{equation*}
\xi_{k}^{2}=\alpha^{2}\xi_{k-1}^{2}+(1-\alpha^{2})\sigma_{k}^{2}.
\end{equation*}
The expectation of the log evidence estimates satisfies
\begin{align*}
\mathbb{E}[\log  w_{\textup{mar}}(x_{K})]&=\frac{1}{2}\log\left(2\pi\xi_{K}^{2}\right)+\frac{1}{2}\left(\frac{1}{\xi_{K}^{2}}-\frac{1}{\sigma_{K}^{2}}\right)\xi_{K}^{2},\\
\mathbb{E}[\log w_{\textup{ais}}(x_{0:K})]&=\frac{1}{2}\log(2\pi\sigma_{0}^{2})+\frac{\left(\beta_{K}-1\right)}{2}\sum_{k=1}^{K}\frac{\xi_{k-1}^{2}}{\sigma_{k}^{2}},%
\end{align*}
while their variance is given by 
\begin{align*}
\textup{var}[\log  w_{\textup{mar}}(x_{K})]&=\frac{1}{2}\left(\frac{1}{\xi_{K}^{2}}-\frac{1}{\sigma_{K}^{2}}\right)^{2}\xi_{K}^{4},\\
\textup{var}[\log w_{\textup{ais}}(x_{0:K})] & =\left(\beta_{K}-1\right)^{2} \bigg( \sum_{k=1}^{K}\frac{\xi_{k-1}^{4}}{2 \sigma_{k}^{4}}+\sum_{K\geq l>k\geq1}^{K}\frac{\alpha^{2(l-k)}\xi_{k-1}^{4}}{\sigma_{k}^{2}\sigma_{l}^{2}}\bigg).
\end{align*}
\end{prop}

\begin{figure}
    \centering
    \includegraphics[width=.99\textwidth]{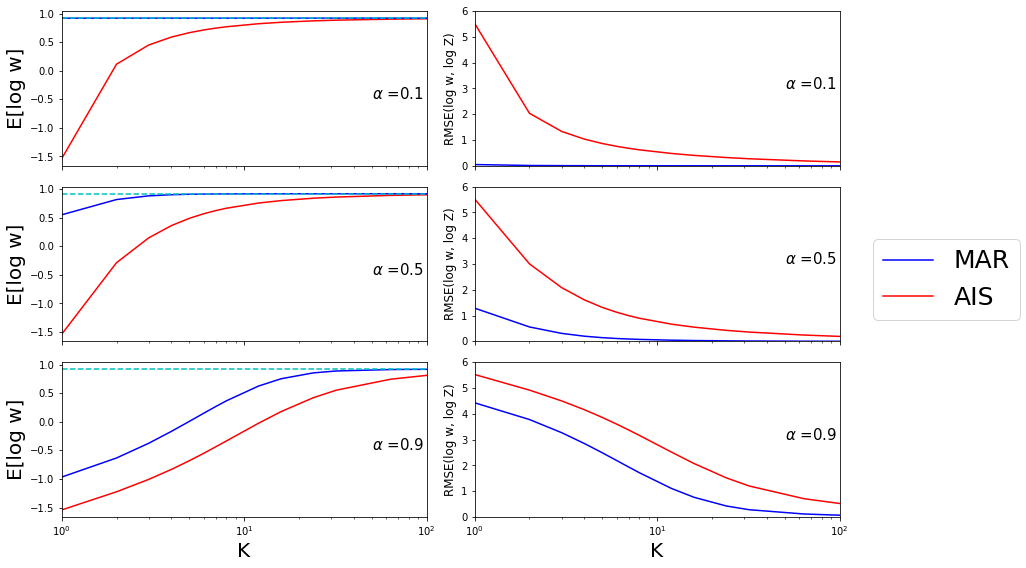}
    \caption{Plot of analytic results from Section \ref{sev:varcalcul}. The left column shows the analytic mean $\mathbb{E}[\log w]$ as a function of the number of temperature transitions $K$ for different values of the mixing parameter $\alpha \in [0, 1]$. $\alpha=0$ corresponds to perfect mixing while $\alpha=1$ corresponds to no mixing. MAR denotes the optimal importance weight, where AIS shows the one from Annealed Importance Sampling. Both estimators tend to the correct value $\log Z$ (shown as a Cyan line) as $K$ becomes large but MAR does so faster. The right column shows the same plots but for the root mean squared error (RMSE) of $\log w$ treated as an estimator of $\log Z$. It is computed as the root of the bias squared plus the variance, i.e $\sqrt{(\mathbb{E}[\log w] - \log Z)^2 + \text{Var}[\log w] }$. The RMSE of both estimators tends to zero in both cases as $K$ becomes large but MAR does so faster.}
    \label{fig:my_label}
\end{figure}

\begin{proof}
The proposal is given by $x_{0}\sim\mathcal{N}(0,\sigma_{0}^{2})$
and
\[
x_{k}=\alpha x_{k-1}+\sqrt{1-\alpha^{2}}\sigma_{k}\epsilon_{k},\quad\quad\epsilon_{k}\overset{\textup{i.i.d.}}{\sim}\mathcal{N}(0,1).
\]
So marginally, under $Q$, we have $x_{k}\sim\mathcal{N}(0,\xi_{k}^{2})$
where $(\xi_{k})_{k\geq 0}$ satisfies the recursion stated above.

We are first looking at the optimal log-estimate of $Z$
which is given by 
\begin{align*}
\log  w_{\textup{mar}}(x_{K}) & =\log\gamma_{K}(x_{K})-\log q_{K}(x_{K})\\
 & =\frac{1}{2}\log\left(2\pi\xi_{K}^{2}\right)+\frac{1}{2}\left(\frac{1}{\xi_{K}^{2}}-\frac{1}{\sigma_{K}^{2}}\right)x_{K}^{2}
\end{align*}
so we have 
\[
\mathbb{E}[\log  w_{\textup{mar}}(x_{K})]=\frac{1}{2}\log\left(2\pi\xi_{K}^{2}\right)+\frac{1}{2}\left(\frac{1}{\xi_{K}^{2}}-\frac{1}{\sigma_{K}^{2}}\right)\xi_{K}^{2}
\]
and, using $\textup{var}[x^{2}]=2\sigma^{4}$ for $x\sim\mathcal{N}(0,\sigma^{2})$, we
obtain 
\[
\textup{var}[\log  w_{\textup{mar}}(x_{K})]=\frac{1}{2}\left(\frac{1}{\xi_{K}^{2}}-\frac{1}{\sigma_{K}^{2}}\right)^{2}\xi_{K}^{4}.
\]

We are now looking at the AIS log-estimate of $Z$ which is given
by
\begin{align*}
\log w_{\textup{ais}}(x_{0:K}) & =\sum_{k=1}^{K}\log(\gamma_{k}(x_{k-1}))-\log(\gamma_{k-1}(x_{k-1}))\\
 & =\frac{1}{2}\log(2\pi\sigma_{0}^{2})+\frac{\left(\beta_{K}-1\right)}{2}\sum_{k=1}^{K}\frac{x_{k-1}^{2}}{\sigma_{k}^{2}},
\end{align*}
the term $\frac{1}{2}\log(2\pi\sigma_{0}^{2})$ coming from the fact
that we consider $\gamma_{0}=\pi_{0}$. It follows that
\begin{align*}
\mathbb{E}[\log w_{\textup{ais}}(x_{0:K})]=\frac{1}{2}\log(2\pi\sigma_{0}^{2})+ & \frac{\left(\beta_{K}-1\right)}{2}\sum_{k=1}^{K}\frac{\xi_{k-1}^{2}}{\sigma_{k}^{2}}.
\end{align*}
The variance is given by
\begin{align*}
\textup{var}[\log w_{\textup{ais}}(x_{0:K})] & =\frac{\left(\beta_{K}-1\right)^{2}}{4}\textup{var}\left[\sum_{k=1}^{K}\frac{x_{k-1}^{2}}{\sigma_{k}^{2}}\right]\\
 & =\frac{\left(\beta_{K}-1\right)^{2}}{4}\sum_{k=1}^{K}\frac{1}{\sigma_{k}^{4}}\textup{var}\left[x_{k-1}^{2}\right]+2\frac{\left(\beta_{K}-1\right)^{2}}{4}\sum_{K\geq l>k\geq1}^{K}\frac{1}{\sigma_{k}^{2}\sigma_{l}^{2}}\textup{cov}\left[x_{k-1}^{2},x_{l-1}^{2}\right].
\end{align*}
Now, using again  $\textup{var}[x^2]=2\sigma^{4}$ for $x\sim\mathcal{N}(0,\sigma^{2})$, we have $\textup{var}\left[x_{k-1}^{2}\right]=2\xi_{k-1}^{4}$.
To compute $\textup{cov}\left[x_{k-1}^{2},x_{l-1}^{2}\right]$, we use the fact that one can easily check from the form of the forward transitions that $(x_{k-1},x_{l-1})$
satisfies for $l>k$
\[
\textup{cov}(x_{k-1},x_{l-1})=\alpha^{l-k}\xi_{k-1}^{2}.
\]

So we have in distribution for $z\sim\mathcal{N}(0,1)$ independent
of $x_{k-1}$
\[
(x_{k-1},x_{l-1})=(x_{k-1},\alpha^{l-k}x_{k-1}+\sqrt{\xi_{l-1}^{2}-\alpha^{2\left(l-k\right)}\xi_{k-1}^{2}}z).
\]
Thus it follows that 
\begin{align*}
x_{k-1}^{2}x_{l-1}^{2} & =x_{k-1}^{2}\left(\alpha^{l-k}x_{k-1}+\sqrt{\xi_{l-1}^{2}-\alpha^{2\left(l-k\right)}\xi_{k-1}^{2}}z\right)^{2}\\
 & =x_{k-1}^{2}\left(\alpha^{2(l-k)}x_{k-1}^{2}+(\xi_{l-1}^{2}-\alpha^{2\left(l-k\right)}\xi_{k-1}^{2})z^{2}+2\alpha^{l-k}\sqrt{\xi_{l-1}^{2}-\alpha^{2\left(l-k\right)}\xi_{k-1}^{2}}x_{k-1}z\right)\\
 & =\alpha^{2(l-k)}x_{k-1}^{4}+(\xi_{l-1}^{2}-\alpha^{2\left(l-k\right)}\xi_{k-1}^{2})z^{2}x_{k-1}^{2}+2\alpha^{l-k}\sqrt{\xi_{l-1}^{2}-\alpha^{2\left(l-k\right)}\xi_{k-1}^{2}}x_{k-1}^{3}z.
\end{align*}
Hence, we obtain
\begin{align*}
\textup{cov}\left[x_{k-1}^{2},x_{l-1}^{2}\right] & =\mathbb{E}\left[x_{k-1}^{2}x_{l-1}^{2}\right]-\mathbb{E}\left[x_{k-1}^{2}\right]\mathbb{E}\left[x_{l-1}^{2}\right]\\
 & =3\alpha^{2(l-k)}\xi_{k-1}^{4}+(\xi_{l-1}^{2}-\alpha^{2\left(l-k\right)}\xi_{k-1}^{2})\xi_{k-1}^{2}-\xi_{k-1}^{2}\xi_{l-1}^{2}\\
 & =2\alpha^{2(l-k)}\xi_{k-1}^{4}.
\end{align*}
This finally yields
\begin{align*}
\textup{var}[\log w_{\textup{ais}}(x_{0:K})] & =\frac{\left(\beta_{K}-1\right)^{2}}{4}\sum_{k=1}^{K}\frac{1}{\sigma_{k}^{4}}\textup{var}\left[x_{k-1}^{2}\right]+2\frac{\left(\beta_{K}-1\right)^{2}}{4}\sum_{K\geq l>k\geq1}^{K}\frac{1}{\sigma_{k}^{2}\sigma_{l}^{2}}\textup{cov}\left[x_{k-1}^{2},x_{l-1}^{2}\right]\\
 & =\frac{\left(\beta_{K}-1\right)^{2}}{2}\sum_{k=1}^{K}\frac{\xi_{k-1}^{4}}{\sigma_{k}^{4}}+\left(\beta_{K}-1\right)^{2}\sum_{K\geq l>k\geq1}^{K}\frac{\alpha^{2(l-k)}\xi_{k-1}^{4}}{\sigma_{k}^{2}\sigma_{l}^{2}},\\
\end{align*}
as required.
\end{proof}
We now illustrate these results in Figure \ref{fig:my_label} by plotting the expectation and the root mean square error of $\log  w_{\textup{mar}}$ and $\log  w_{\textup{ais}}$ for various $\alpha$ and $K$.

\section{Proof of Propositions}
\subsection{Proof of Proposition \ref{prop:optimalextentedtarget}}
\begin{proof}
The chain rule for the Kullback--Leibler divergence $\KL(Q||P)$ yields
\begin{equation}\label{eq:KLdecompo}
    \KL(Q||P)=\KL(q_K||\pi)+\E_{q_K}\Big[\KL(Q(\cdot|x_K)||P(\cdot|x_K))\Big],
\end{equation}
where, from (\ref{eq:AISproposal}) and (\ref{eq:AIStarget}), the conditional distributions of $x_{0:K-1}$ given $x_K$ are equal to
\begin{equation}\label{eq:backwarddecompoPQ}
    Q(x_{0:K-1}|x_K)=\prod_{k=0}^{K-1}B^{\textup{opt}}_k(x_k|x_{k+1}),\quad\quad
    P(x_{0:K-1}|x_K)=\prod_{k=0}^{K-1}B_k(x_k|x_{k+1}),
\end{equation}
The expression for $Q$ above follows directly from its time-reversed decomposition; i.e.
\begin{align}
    Q(x_{0:K})=q_K(x_K)\prod_{k=0}^{K-1} \frac{q_{k}(x_{k})F_{k+1}(x_{k+1}|x_k)}{q_{k+1}(x_{k+1})}=q_K(x_K)\prod_{k=0}^{K-1}B^{\textup{opt}}_k(x_k|x_{k+1}),
\end{align}
where we recall that $q_0(x_0)=\pi_0(x_0)$. It thus follows directly from (\ref{eq:KLdecompo}) and (\ref{eq:backwarddecompoPQ}) that the backward transition kernels $(B_k)_{k=0}^{K-1}$ minimizing $\KL(Q||P)$ are $(B^{\textup{opt}}_k)_{k=0}^{K-1}$ as this implies $P(x_{0:K-1}|x_K)=Q(x_{0:K-1}|x_K)$.

The variance decomposition formula yields for all $P$
\begin{align}
     \textup{var}_Q[w(x_{0:K})]&=\textup{var}_{q_K}[\E_{Q(\cdot|x_K)}[w(x_{0:K})]]+\E_{q_K}[\textup{var}_{Q(\cdot|x_K)}[w(x_{0:K})]] \nonumber\\
     &=\textup{var}_{q_K}[w_{\textup{mar}}(x_{K})]+\E_{q_K}[\textup{var}_{Q(\cdot|x_K)}[w(x_{0:K})]] \nonumber\\
     &\geq \textup{var}_{q_K}[w_{\textup{mar}}(x_{K})].\nonumber
\end{align}
By direct calculations, we also have $ w_{\textup{mar}}(x_K)=\Gamma^{\textup{opt}}(x_{0:K})/Q(x_{0:K})$ so $P^\textup{opt}$ minimizes the variance of the evidence estimate.
\end{proof}

\subsection{Proof of Proposition \ref{prop:ELBOmixed}}\label{appendixA2}
We establish first here Proposition \ref{prop:ELBOcontinuoustime} and  Proposition \ref{prop:ELBOdiscretetime}. Both results can then be easily combined to obtain Proposition \ref{prop:ELBOmixed}.

\begin{prop}\label{prop:ELBOcontinuoustime} Under regularity conditions, one has 
\begin{align}\label{eq:differencescores}
    \KL(\mathcal{Q}||\mathcal{P}_\theta)&= \E_{\mathcal{Q}}\Big[\int_0^T||s_{\theta}(t,x_t)-\nabla \log q_t(x_t)||^2\rmd t\Big]+C_1\\
    &=\sum_{k=1}^{K}\int_{t_{k-1}}^{t_{k}}\E_\mathcal{Q}\left[||s_{\theta}(t,x_{t})-\nabla \log q_{t|t_{k-1}}(x_{t}|x_{t_{k-1}})||^{2}\right]\rmd t+C_2,\label{eq:scorematching}
\end{align}
for constants $C_1,C_2$ independent of $\theta$, where $t_k=k \delta$ with $K=T/\delta$ and $q_{t|s}(x'|x)$ is the density of $X_{t}=x'$ given $X_s=x$ under $\mathcal{Q}$.

\end{prop}

To establish (\ref{eq:differencescores}), we follow arguments similar to \cite[Theorem 2]{song2021maximum}. The loss (\ref{eq:scorematching}) we then consider differs from the one uses in the score-based generative modeling literature. This is because, contrary to the Ornstein--Ulhenbeck process used for SGM, the transition density $q_{t'|t}(x'|x)$ of the forward diffusion (\ref{eq:nonhomogeneousLangevin}) is not available in closed-form and can only be approximated reliably when $t'-t$ is small. Practically, to obtain a tractable criterion, we need to first approximate the integrals in (\ref{eq:scorematching}) by the rectangular rule. We also discretize the Langevin dynamics using an Euler--Maruyama scheme; i.e. we use an approximation $Q$ of $\mathcal{Q}$ based on the ULA kernel $F_k(x_k|x_{k-1})=\mathcal{N}(x_k;x_{k-1}+\delta \nabla \log \pi_{k}(x_{k-1});2\delta I)$ approximating $q_{t_k|t_{k-1}}(x'|x)$. We recall here that we slightly abuse notation by writing $\pi_{t_k}=\pi_{k \delta}$ as $\pi_k$. We thus finally obtain a loss 
\begin{equation}\label{eq:approximatescorematchingloss}
 \mathcal{L}(\theta)= \delta \sum_{k=1}^{K} \E_Q\left[||s_{\theta}(k,x_{k})-\nabla \log F_k(x_{k}|x_{k-1})||^{2}\right].
\end{equation}

\begin{proof}
We assume here sufficient regularity conditions ensuring that the SDEs given below admit a unique solution, their corresponding time-reversed diffusions are well-defined and Girsanov theorem applies; see e.g. \cite[Appendix A]{song2021maximum}.

By the chain rule for KL divergence, one has 
\begin{equation}
    \KL(\mathcal{Q}||\mathcal{P}_\theta)=\KL(q_T||\pi)+\E_{q_T}\Big[\KL(\mathcal{Q}(\cdot|x_T)||\mathcal{P}_\theta(\cdot|x_T))\Big] 
\end{equation}
where $\mathcal{Q}(\cdot|x_T)$ and $\mathcal{P}_\theta(\cdot|x_T)$ are the path measures induced by 
\begin{equation}\label{eq:Langevinreversalpinned}
    \rmd \bar{x}_t=\big\{-\nabla \log \pi_{T-t}(\bar{x}_t)+2 \nabla \log q_{T-t}(\bar{x}_t)\big\} \rmd t+\sqrt{2} \rmd \bar{B}_t,\quad\quad \bar{x}_0=x_T,
\end{equation}
and 
\begin{equation}\label{eq:Langevinreversalapproxpinned}
    \rmd \bar{x}_t=\big\{-\nabla \log \pi_{T-t}(\bar{x}_t)+2 \nabla \log s_\theta(T-t,\bar{x}_t)\big\} \rmd t+\sqrt{2} \rmd \bar{B}_t,\quad\quad \bar{x}_0=x_T.
\end{equation}
We now use Girsanov theorem (see e.g. \citep[Section 10.3]{klebaner2012introduction} and \citep{Opper2019}) to compute the Radon--Nikodym derivative $\rmd {\mathcal{Q}}(\cdot|x_T)/\rmd \mathcal{P}_\theta(\cdot|x_T)$ so that
\begin{align*}
    &\E_{q_T}\Big[\KL(\mathcal{Q}(\cdot|x_T)||\mathcal{P}_\theta(\cdot|x_T))\Big]\\
    =&-\E_{\mathcal{Q}}\Big[\log \frac{\rmd \mathcal{P}_\theta(\cdot|x_T)}{\rmd {\mathcal{Q}}(\cdot|x_T)}\Big]\\
    =&\E_{\mathcal{Q}}\Big[\sqrt{2} \int_0^T (\nabla \log q_{T-t}(\bar{x}_t)-s_\theta(T-t,\bar{x}_t))\rmd \bar{B}_t+\int_0^T||\nabla \log q_{T-t}(\bar{x}_t)-s_\theta(T-t,\bar{x}_t)||^2\rmd t\Big]\\
    =&\E_{\mathcal{Q}}\Big[\int_0^T||\nabla \log q_t(x_t)-s_\theta(t,x_t)||^2\rmd t\Big],
\end{align*}
as $\E_{\mathcal{Q}}\Big[\int_0^T f_t(\bar{x}_t)\rmd \bar{B}_t\Big]=0$ for any function $f_t$.

As in \citep{Heng2021diffusionbridges} in a different context, we can write for
any partition of $[0,T]$ defined by $t_{0}=0<t_{1}<\cdots<t_{K-1}<t_{K}=T$
\begin{align*}
    \E_{\mathcal{Q}}\Big[\int_0^T||\nabla \log q_t(x_t)-s_\theta(t,x_t)||^2\rmd t\Big]&=\int_0^T  \int ||\nabla \log q_t(x)-s_\theta(t,x)||^2 q_t(x) \rmd x \rmd t\\
    &=\sum_{k=1}^{K}\int_{t_{k-1}}^{t_{k}}\int||\nabla \log q_t(x)-s_\theta(t,x)||^2 q_t(x) \rmd x \rmd t
\end{align*}
where, for a constant $c$ independent of $\theta$, we have
\begin{align*}
 & \int_{t_{k-1}}^{t_{k}}\int||\nabla \log q_t(x)-s_\theta(t,x)||^2 q_t(x) \rmd x \rmd t \\
= & \int_{t_{k-1}}^{t_{k}}\int\left\{ ||\nabla \log q_t(x)||^{2}+||s_\theta(t,x)||^{2}-2s_\theta(t,x)^{\textup{T}}\nabla \log\ q_{t}(x)\right\} q_{t}(x)\rmd x\rmd t\\
= & \int_{t_{k-1}}^{t_{k}}\int\left\{||s_\theta(t,x)||^{2}-2s_\theta(t,x)^{\textup{T}}\nabla \log\ q_{t}(x)\right\} q_{t}(x)\rmd x\rmd t +c.\\
\end{align*}
Now we have 
\begin{equation}
\int_{t_{k-1}}^{t_{k}}\int s_\theta(t,x)^{\textup{T}}\nabla \log q_{t}(x) q_{t}(x)\rmd x\rmd t = \int_{t_{k-1}}^{t_{k}}\int s_\theta(t,x)^{\textup{T}}\nabla  q_{t}(x) \rmd x\rmd t
\end{equation}
where, using Chapman-Kolmogorov, $q_t$ satisfies
\begin{equation}
q_{t}(x)=\int q_{t_{k-1}}(x_{t_{k-1}}) q_{t|t_{k-1}}(x|x_{t_{k-1}}) \rmd x_{t_{k-1}}.
\end{equation}
It follows that 
\begin{equation}
\nabla q_{t}(x)=\int q_{t_{k-1}}(x_{t_{k-1}}) \nabla q_{t|t_{k-1}}(x|x_{t_{k-1}}) \rmd x_{t_{k-1}}.
\end{equation}

Hence, we have
\begin{align*}
 & \int_{t_{k-1}}^{t_{k}}\int s_\theta(t,x)^{\textup{T}}\nabla  q_{t}(x) \rmd x\rmd t\\
= & \int_{t_{k-1}}^{t_{k}}\int\int s_{\theta}(t,x)^{\textup{T}} 
\nabla \log q_{t|t_{k-1}}(x|x_{t_{k-1}}) q_{t_{k-1}}(x_{t_{k-1}})q_{t|t_{k-1}}(x|x_{t_{k-1}})
\rmd x_{t_{k-1}}\rmd x \rmd t
\end{align*}
so minimizing $\E_{q_T}\Big[\KL(\mathcal{Q}(\cdot|x_T)||\mathcal{P}_\theta(\cdot|x_T))\Big]$ w.r.t.
$\theta$ is equivalent to minimize
\begin{align*}
&\sum_{k=1}^{K}\int_{t_{k-1}}^{t_{k}}\int\int ||s_\theta(t,x)||^{2} q_{t_{k-1}}(x_{t_{k-1}}) q_{t|t_{k-1}}(x|x_{t_{k-1}})\rmd x_{t_{k-1}}\rmd x\rmd t \\
-2&\sum_{k=1}^{K}\int_{t_{k-1}}^{t_{k}}\int\int s_{\theta}(t,x)^{\textup{T}} 
\nabla \log q_{t|t_{k-1}}(x|x_{t_{k-1}}) q_{t_{k-1}}(x_{t_{k-1}}) q_{t|t_{k-1}}(x|x_{t_{k-1}})
\rmd x_{t_{k-1}}\rmd x \rmd t\\
=&\sum_{k=1}^{K}\int_{t_{k-1}}^{t_{k}}\int\int ||s_\theta(t,x)-\nabla \log q_{t|t_{k-1}}(x|x_{t_{k-1}})||^2 q_{t_{k-1}}(x_{t_{k-1}})q_{t|t_{k-1}}(x|x_{t_{k-1}})\rmd x_{t_{k-1}}\rmd x \rmd t+C
\end{align*}
where $C$ is independent of $\theta$. Hence, this is equivalent to minimizing (\ref{eq:scorematching}).
\end{proof}

We now establish results about the discrete-time Kullback--Leibler divergence $\KL(Q||P_\theta)$. First note that \begin{align}
    \KL(Q||P_\theta)&=\E_Q\Big[\log \frac{Q(x_{0:K})}{P_\theta(x_{0:K})}\Big] \nonumber\\
    &=\E_Q\bigg[\log \frac{\pi_0(x_0)\prod_{k=0}^{K-1} F_{k+1}(x_{k+1}|x_{k})}{\pi(x_K)\prod_{k=0}^{K-1} B^{\theta}_k(x_k|x_{k+1})}\bigg] \nonumber\\
    &=-\sum_{k=0}^{K-1} \E_Q\big[\log B^{\theta}_k(x_k|x_{k+1})\big]+C_1,\label{eq:ELBOdiscrete}
\end{align}
where, as $B^{\theta}_k(x_k|x_{k+1})=\mathcal{N}(x_k;x_{k+1}-\delta \nabla \log \pi_{k+1}(x_{k+1})+2 \delta s_\theta(k+1,x_{k+1}),2\delta I)$, one has 
\begin{align}
    -\log B^{\theta}_k(x_k|x_{k+1})&=\frac{1}{4\delta}||x_k-x_{k+1}+\delta \nabla \log \pi_{k+1}(x_{k+1})-2\delta s_{\theta}(k+1,x_{k+1})||^2+C_2 \nonumber\\
    &=\delta \Big\|s_{\theta}(k+1,x_{k+1})-\frac{1}{2\delta}(x_k-x_{k+1}+\delta \nabla \log \pi_{k+1}(x_{k+1}))\Big\|^2+C_2 \label{eq:decompoB}\\
    &\approx \delta \Big\|s_{\theta}(k+1,x_{k+1})-\frac{1}{2\delta}(x_k-x_{k+1}+\delta \nabla \log \pi_{{k+1}}(x_{k}))\Big\|^2+C_2 \nonumber\\
    &= \delta \Big\|s_{\theta}(k+1,x_{k+1})- \nabla \log F_{k+1}(x_{k+1}|x_k)\Big\|^2+C_2 \label{eq:decompoBapprox},
\end{align}
where we have used $\pi_{k+1}(x_{k+1})\approx \pi_{k+1}(x_k)$ for $\delta \ll 1$. The sum over $k=0,...,K-1$ of the first terms on the r.h.s. of (\ref{eq:decompoB}) are equal to the loss $\mathcal{L}(\theta)$ defined in (\ref{eq:approximatescorematchingloss}). More rigorously, we can prove the following result.

\begin{assum}\label{assum:gradientLip}
There exists $L\leq \infty$ such that for all $k$ and $x,x' \in \mathbb{R}^d$ 
\begin{equation}
   \Big\| \nabla \log \pi_k(x)-\nabla \log \pi_k(x') \Big\|\leq L \Big\|x-x' \Big\|.
\end{equation}
\end{assum}

\begin{assum}\label{assum:moments}
There exists $C\leq \infty$ such that
\begin{equation}\label{eq:bound1}
\limsup_K \max_{k=0,...,K-1} \E_{Q_K}\left[ \Big\|\nabla \log \pi_{k+1}(x_{k})\Big\|^2\right] \leq C
\end{equation}
and for any $\theta$
\begin{equation}\label{eq:bound2}
\limsup_K \max_{k=0,...,K-1}  \E_{Q_K}\left[ \Big\| \nabla_{\theta} s_{\theta}(k+1,x_{k+1})\Big\|^2\right] \leq C,
\end{equation}
where we have emphasized here notationally that $Q$ is a function of $K$.
\end{assum}

\begin{prop}\label{prop:ELBOdiscretetime} Under Assumptions \ref{assum:gradientLip}-\ref{assum:moments}, the gradient of the Kullback--Leibler divergence $\KL(Q||P_\theta)$ satisfies 
\begin{equation}\label{eq:K}
    \nabla \KL(Q||P_\theta)= \nabla \mathcal{L}(\theta)+\epsilon(\theta),
\end{equation}
for $\mathcal{L}(\theta)$ defined in (\ref{eq:approximatescorematchingloss}) and a function $\epsilon$ satisfying $\lim_{K\rightarrow \infty}\epsilon(\theta)=0$.
\end{prop}

\begin{proof}
In the rest of the proof, all the expectations are taken w.r.t. $Q$ unless mentioned otherwise and we drop it from the notations for simplicity. However as we take gradients w.r.t. to both $x$ and $\theta$, this is indicated notationally to avoid confusion. We also assume that $\theta$ is a scalar in the proof, the extension to the multivariate case is straightforward. 

Using (\ref{eq:ELBOdiscrete}), we have  
\begin{align}
    \nabla_{\theta} \KL(Q||P_\theta)&=-\sum_{k=0}^{K-1} \E\big[ \nabla_{\theta} \log B^{\theta}_k(x_k|x_{k+1})\big],\label{eq:gradELBOdiscrete}
\end{align}
where, from (\ref{eq:decompoB}), one has 
\begin{align}
    -& \nabla_{\theta} \log B^{\theta}_k(x_k|x_{k+1})\\
    =&\delta \nabla_{\theta}  \Big\|s_{\theta}(k+1,x_{k+1})-\frac{1}{2\delta}(x_k-x_{k+1}+\delta \nabla_x \log \pi_{{k+1}}(x_{k+1}))\Big\|^2 \nonumber\\
    =&2 \delta \nabla_{\theta} s_{\theta}(k+1,x_{k+1})^{\textup{T}} (s_{\theta}(k+1,x_{k+1})-\frac{1}{2\delta}(x_k-x_{k+1}+\delta \nabla_x \log \pi_{{k+1}}(x_{k+1}))).
\end{align}
We also have 
\begin{align}
   \nabla_{\theta} \mathcal{L}(\theta)&=\delta \sum_{k=0}^{K-1}  \E\left[ \nabla_{\theta} \Big\|s_{\theta}(k,x_k)-\nabla_x \log F_k(x_{k}|x_{k-1})\Big\|^{2}\right],\label{eq:gradELBOdiscrete}
\end{align}
where 
\begin{align}
    &\delta  \nabla_{\theta}  \Big\|s_{\theta}(k,x_k)-\nabla_x \log F_k(x_{k}|x_{k-1})\Big\|^2 \nonumber\\
    =&\delta  \nabla_{\theta}  \Big\|s_{\theta}(k,x_k)-\frac{1}{2\delta}(x_k-x_{k+1}+\delta \nabla_x \log \pi_{{k+1}}(x_{k}))\Big\|^2 \nonumber\\
    =&2 \delta  \nabla_{\theta} s_{\theta}(k+1,x_{k+1})^{\textup{T}} (s_{\theta}(k+1,x_{k+1})-\frac{1}{2\delta}(x_k-x_{k+1}+\delta \nabla_x \log \pi_{{k+1}}(x_{k}))).
\end{align}

So we obtain by using (\ref{eq:ELBOdiscrete}) and (\ref{eq:decompoB})
\begin{equation}
      \nabla_{\theta} \KL(Q||P_\theta)= \nabla_{\theta} \mathcal{L}(\theta)+ \epsilon(\theta),
\end{equation}
for 
\begin{equation}
    \epsilon(\theta)=2\delta \E\left[\sum_{k=0}^{K-1}  \nabla_{\theta} s_{\theta}(k+1,x_{k+1})^{\textup{T}} (\nabla_x \log \pi_{{k+1}}(x_{k})-\nabla_x \log \pi_{{k+1}}(x_{k+1})) \right].
\end{equation}

Hence we have 
\begin{align}
      |\epsilon(\theta)| &\leq 2\delta \sum_{k=0}^{K-1} \E\left[ | \nabla_{\theta} s_{\theta}(k+1,x_{k+1})^{\textup{T}} (\nabla_x \log \pi_{{k+1}}(x_{k})-\nabla_x \log \pi_{{k+1}}(x_{k+1}))| \right] \nonumber \\
      &\leq 2\delta \sum_{k=0}^{K-1}  \E\left[\Big\| \nabla_{\theta} s_{\theta}(k+1,x_{k+1})\Big\|^2 \right]^{1/2}\E\left[\Big\|\nabla_x \log \pi_{k+1}(x_{k}) -\nabla_x \log \pi_{k+1}(x_{k+1}) \Big\|^2\right]^{1/2}\label{eq:boundepsilon}
 \end{align}

From Assumption \ref{assum:gradientLip}, we have 
\begin{align}
    \E\left[\Big\|\nabla_x \log \pi_{k+1}(x_{k}) -\nabla_x \log \pi_{k+1}(x_{k+1}) \Big\|^2\right]
    &\leq L^2 \E\left[\Big\|x_{k+1} -x_{k} \Big\|^2\right] \nonumber\\
    &\leq 2 L^2 \delta \E\left[\delta \Big\|\nabla_x \log \pi_{k+1}(x_{k})\Big\|^2 + 2 M \right] \label{eq:Lipgradbound},
\end{align}
where $M=\mathbb{E}_{Z\sim \mathcal{N}(0,I)}[||Z||^2]$ as $x_{k+1}=x_k+\delta \nabla_x \log \pi_{k+1}(x_{k})+\sqrt{2\delta}Z$ under $Q$.
Now using Assumption \ref{assum:moments}, it follows from (\ref{eq:boundepsilon}), (\ref{eq:Lipgradbound}), (\ref{eq:bound1}), (\ref{eq:bound2}) and $K=O(1/\delta)$ that $\epsilon(\theta)=O(\sqrt{\delta})$. The result follows.

\end{proof}

\subsection{Proof of Proposition \ref{prop:ELBOunderdampedCT}}
We also assume here sufficient regularity conditions ensuring that the SDEs given below admit a unique solution, their corresponding time-reversed diffusions are well-defined and Girsanov theorem applies.
The proof is very similar to the proof of the first part of Proposition \ref{prop:ELBOmixed} (i.e. Proposition \ref{prop:ELBOcontinuoustime} in Appendix \ref{appendixA2}). 

To start with, we use again the chain rule for KL divergences, one has 
\begin{equation}\label{eq:chainruleSM}
    \KL(\mathcal{Q}||\mathcal{P}_\theta)=\KL(\eta_T||\pi_T \otimes \mathcal{N}(0,M))+\E_{\eta_T}\Big[\KL(\mathcal{Q}(\cdot|x_T,p_T)||\mathcal{P}_\theta(\cdot|x_T,p_T))\Big] 
\end{equation}
where $\mathcal{Q}(\cdot|x_T,p_T)$ is the path measure induced by 
\begin{align}\label{eq:timereversalnonhomogeneousunderdampedLangevinpositivetimeSM}
\rmd \bar{x}_{t} & =-M^{-1}\bar{p}_{t}\rmd t,\\
\rmd \bar{p}_{t} & =-\nabla \log \pi_{T-t}(\bar{x}_{t})\rmd t+\zeta \bar{p}_{t}\rmd t+2\zeta M\nabla_{\bar{p}_{t}}\log \eta_{T-t}(\bar{x}_{t},\bar{p}_{t})\rmd t+\sqrt{2\zeta}M^{1/2}\rmd \bar{B}_{t}\nonumber
\end{align}
and $\mathcal{P}_\theta(\cdot|x_T,p_T)$ the path measure induced by
\begin{align}\label{eq:approximatetimereversalnonhomogeneousunderdampedLangevinpositivetimeSM}
&\rmd \bar{x}_{t} =-M^{-1}\bar{p}_{t}\rmd t,\\
&\rmd \bar{p}_{t} =-\nabla \log \pi_{T-t}(\bar{x}_{t})\rmd t+\zeta \bar{p}_{t}\rmd t + 2\zeta M s_{\theta}(T-t,\bar{x}_t,\bar{p}_t)+\sqrt{2\zeta}M^{1/2}\rmd \bar{B}_{t},\nonumber
\end{align}
these two diffusions being initialized at $(\bar{x}_{0},\bar{p}_{0})=(x_T,p_T)$.

By now applying a version of Girsanov's theorem that allows for some of the components of the diffusions to be noiseless \cite[Theorem 4]{sottinen2008application}, we obtain
\begin{align}
    &\E_{\eta_T}\Big[\KL(\mathcal{Q}(\cdot|x_T,p_T)||\mathcal{P}_\theta(\cdot|x_T,p_T))\Big] \nonumber\\
    =& \frac{1}{2}\mathbb{E}_{\mathcal{Q}}\Big[\int_{0}^{T}\int ||2\zeta M\nabla_{p}\log\eta_{t}(x_t,p_t)-2\zeta Ms_{\theta}(t,x_t,p_t)||_{\left(2\zeta M\right)^{-1}}^{2}\mathrm{d}t\Big] \nonumber\\
  =& \zeta \mathbb{E}_{\mathcal{Q}}\Big[\int_{0}^{T} ||\nabla_{p}\log\eta_{t}(x_t,p_t)-s_{\theta}(t,x_t,p_t)||_{M}^{2}\mathrm{d}t\Big].\label{eq:girsanovunderdamped}
\end{align}
Now equation (\ref{eq:scorematchingunderdamped1}) follows directly from (\ref{eq:chainruleSM}) and (\ref{eq:girsanovunderdamped}). We are now going to rewrite this loss to make it more tractable. We have for
any partition of $[0,T]$ defined by $t_{0}=0<t_{1}<\cdots<t_{K-1}<t_{K}=T$
\begin{align*}
&\mathbb{E}_{\mathcal{Q}}\Big[\int_{0}^{T} ||\nabla_{p}\log\eta_{t}(x_t,p_t)-s_{\theta}(t,x_t,p_t)||_{M}^{2}\mathrm{d}t\Big]\\
=&\int_{0}^{T} \int ||\nabla_{p}\log\eta_{t}(x,p)-s_{\theta}(t,x,p)||_{M}^{2}~\eta_{t}(x,p)\mathrm{d}x\mathrm{d}p\mathrm{d}t \\ =&\sum_{k=1}^{K}\int_{t_{k-1}}^{t_{k}}\int ||\nabla_{p}\log\eta_{t}(x,p)-s_{\theta}(t,x,p)||_{M}^{2}~\eta_{t}(x,p)\mathrm{d}x\mathrm{d}p\mathrm{d}t
\end{align*}
where
\begin{align*}
 & \int_{t_{k-1}}^{t_{k}}\int ||\nabla_{p}\log\eta_{t}(x,p)-s_{\theta}(t,x,p)||_{M}^{2}\eta_{t}(x,p)\mathrm{d}x\mathrm{d}p\mathrm{d}t\\
= & \int_{t_{k-1}}^{t_{k}}\int\left\{ ||\nabla_{p}\log\eta_{t}(x,p)||^{2}+||s_{\theta}(t,x,p||_{M}^{2}-2s_{\theta}(t,x,p)^{\textup{T}} M\nabla_{p}\log\eta_{t}(x,p)\right\} \eta_{t}(x,p)\mathrm{d}x\mathrm{d}p\mathrm{d}t\\
= &\int_{t_{k-1}}^{t_{k}}\int\left\{ ||s_{\theta}(t,x,p)||_{M}^{2}-2s_{\theta}(t,x,p)^{\textup{T}} M\nabla_{p}\log\eta_{t}(x,p)\right\} \eta_{t}(x,p)\mathrm{d}x\mathrm{d}p\mathrm{d}t+c.
\end{align*}
Now we have
\begin{equation*}
\int_{t_{k-1}}^{t_{k}}\int s_{\theta}(t,x,p)^{\textup{T}}M\nabla_{p}\log\eta_{t}(x,p)\eta_{t}(x,p)\mathrm{d}x\mathrm{d}p\mathrm{d}t=\int_{t_{k-1}}^{t_{k}}\int  s_{\theta}(t,x,p)^{\textup{T}}M\nabla_{p}\eta_{t}(x,p)\mathrm{d}x\mathrm{d}p\mathrm{d}t
\end{equation*}
where, using Chapman-Kolmogorov, we have
\begin{equation*}
\eta_{t}(x,p)=\int\eta_{t_{k-1}}(x_{t_{k-1}},p_{t_{k-1}}) \eta_{t|t_{k-1}}(x,p|x_{t_{k-1}},p_{t_{k-1}})\mathrm{d}x_{t_{k-1}}\mathrm{d}p_{t_{k-1}}.
\end{equation*}
Here $\eta_{t|t_{k-1}}(x,p|x_{t_{k-1}},p_{t_{k-1}})$ denote the transition
density of $(x_{t},p_{t})=(x,p)$ given $(x_{t_{k-1}},p_{t_{k-1}})$
under the forward dynamics (\ref{eq:underdampedLangevin})
so that
\begin{equation*}
\nabla_p \eta_{t}(x,p)=\int\eta_{t_{k-1}}(x_{t_{k-1}},p_{t_{k-1}}) \nabla_p \eta_{t|t_{k-1}}(x,p|x_{t_{k-1}},p_{t_{k-1}})\mathrm{d}x_{t_{k-1}}\mathrm{d}p_{t_{k-1}}.
\end{equation*}
Hence, it follows that
\begin{align*}
 & \int_{t_{k-1}}^{t_{k}} \int s_{\theta}(t,x,p)^{\textup{T}} M\nabla_{p}\eta_{t}(x,p)\mathrm{d}x\mathrm{d}p\mathrm{d}t\\
= & \int_{t_{k-1}}^{t_{k}}\int\int\eta_{t_{k-1}}(x_{t_{k-1}},p_{t_{k-1}})\eta_{t|t_{k-1}}(x,p|x_{t_{k-1}},p_{t_{k-1}}) \times \\
&\quad\quad\quad\quad \times s_{\theta}(t,x,p)^{\textup{T}} M\nabla_{p}\log \eta_{t|t_{k-1}}(x,p|x_{t_{k-1}},p_{t_{k-1}})\mathrm{d}x\mathrm{d}p\mathrm{d}x_{t_{k-1}}\mathrm{d}p_{t_{k-1}}\mathrm{d}t
\end{align*}
so $\zeta^{-1}\KL(\mathcal{Q}||\mathcal{P}_\theta)$ is equal up to an additive constant $c$ independent of $\theta$ to 
\begin{align*}
 &\sum_{k=1}^{K}\int_{t_{k-1}}^{t_{k}}\int\int s_{\theta}(t,x,p)^{\textup{T}}Ms_{\theta}(t,x,p)\eta_{t_{k-1}}(x_{t_{k-1}},p_{t_{k-1}})\eta_{t|t_{k-1}}(x,p|x_{t_{k-1}},p_{t_{k-1}})\mathrm{d}x\mathrm{d}p\mathrm{d}x_{t_{k-1}}\mathrm{d}p_{t_{k-1}}\mathrm{d}t\\
 -&2\sum_{k=1}^{K}\int_{t_{k-1}}^{t_{k}}\int\int\eta_{t_{k-1}}(x_{t_{k-1}},p_{t_{k-1}})\eta_{t|t_{k-1}}(x,p|x_{t_{k-1}},p_{t_{k-1}})s_{\theta}(t,x,p)^{\textup{T}} M \times \\
 &\quad\quad\quad\quad\quad \times \nabla_{p}\log \eta_{t|t_{k-1}}(x,p|x_{t_{k-1}},p_{t_{k-1}})\mathrm{d}x\mathrm{d}p\mathrm{d}x_{t_{k-1}}\mathrm{d}p_{t_{k-1}}\mathrm{d}t
\end{align*}
\begin{align*}
=&\sum_{k=1}^{K}\int_{t_{k-1}}^{t_{k}}\int\int||s_{\theta}(t,x,p)-\nabla_{p}\log \eta_{t|t_{k-1}}(x,p|x_{t_{k-1}},p_{t_{k-1}})||_{M}^{2}\eta_{t_{k-1}}(x_{t_{k-1}},p_{t_{k-1}})\times \\
&\quad\quad\quad\quad\quad \times \eta_{t|t_{k-1}}(x,p|x_{t_{k-1}},p_{t_{k-1}})\mathrm{d}x\mathrm{d}p\mathrm{d}x_{t_{k-1}}\mathrm{d}p_{t_{k-1}}\mathrm{d}t+c
\end{align*}
Hence, finally we obtain as required
\begin{align}
\KL(\mathcal{Q}||\mathcal{P}_\theta)= & \zeta \sum_{k=1}^{K}\int_{t_{k-1}}^{t_{k}}\mathbb{E}\left[||s_{\theta}(t,x_{t},p_{t})-\nabla_{p_{t}}\log \eta_{t|t_{k-1}}(x_{t},p_{t}|x_{t_{k-1}},p_{t_{k-1}})||_{M}^{2}\right]\mathrm{d}t+C_2.
\end{align}

\section{Details of the reverse integrator for the Hamiltonian variant}\label{app:reverseintegratorHamiltonian}

We take equation (\ref{eq:approximatetimereversalnonhomogeneousunderdampedLangevinpositivetime}) and rewrite in the following form.
\begin{align}\label{eq:approximatetimereversalnonhomogeneousunderdampedLangevinpositivetimepartition}
&\rmd \bar{x}_{t} =-M^{-1}\bar{p}_{t}\rmd t,\\
&\rmd \bar{p}_{t} =-\nabla \log \pi_{T-t}(\bar{x}_{t})\rmd t+2\zeta[M s_{\theta}(T-t,\bar{x}_t,\bar{p}_t)+ \bar{p}_{t}]\rmd t+[-\zeta \bar{p}_{t}\rmd t+\sqrt{2\zeta}M^{1/2}\rmd \bar{B}_{t}].\nonumber
\end{align}

As in \citep{Dockhorn2022}, we have rewritten the SDE for $\bar{p}_t$ in three distinct components, the last one corresponding to an Ornstein--Ulhenbeck process. The integrator we propose however differs from \citep{Dockhorn2022} and will rely on a leapfrog component. In details, we integrate successively over the time interval $\bar{x} \in [t'-\delta, t'] \implies x \in [t+\delta,t]$, where $t = T-t'$. This allows us to directly compare random variables with the forward integrator in Section \ref{section:HamiltonianMain}.

First, we integrate
\begin{equation}\label{eq:Hamiltoniantimereversal}
\rmd \bar{x}_{t} =-M^{-1}\bar{p}_{t}\rmd t,\qquad 
\rmd \bar{p}_{t}  =-\nabla \log \pi_{T-t}(\bar{x}_{t})\rmd t   
\end{equation}
by using $(\tilde{x}_{t+\delta},\tilde{p}_{t+\delta})=\Phi^{-1}_t(x_{t+\delta},p_{t+\delta})$ where $\Phi^{-1}_t(x,p)=\Phi_{\textup{flip}} \circ \Phi_t \circ \Phi_{\textup{flip}}(x,p)$ with $\Phi_{\textup{flip}}(x,p)=(x,-p)$. We then integrate 
\begin{equation}\label{eq:scorereversal}
\rmd \bar{x}_{t} =0, \qquad \rmd \bar{p}_{t} =2\zeta[M s_\theta(T-t,\bar{x}_{t},\bar{p}_{t})+ \bar{p}_{t}]\rmd t
\end{equation}
using  $(\hat{x}_t,\hat{p}_t)=(\tilde{x}_{t+\delta},f_{\theta}(t+\delta,\tilde{x}_{t+\delta},\tilde{p}_{t+\delta}))$ for $f_{\theta}(t+\delta,\tilde{x}_{t+\delta},\tilde{p}_{t+\delta})=\tilde{p}_{t+\delta}+2\delta\zeta[M s_\theta(t+\delta,\tilde{x}_{t+\delta},\tilde{p}_{t+\delta})+\tilde{p}_{t+\delta}]$ and finally
\begin{equation}\label{eq:OU}
\rmd \bar{x}_{t} =0, \qquad \rmd \bar{p}_{t}=-\zeta \bar{p}_{t}\rmd t+\sqrt{2\zeta}M^{1/2}\rmd \bar{B}_{t}
\end{equation}
using $(x_t,p_t)=(\hat{x}_t,h \hat{p}_t+\sqrt{1-h^2}\epsilon)$ with $\epsilon \sim \mathcal{N}(0,M)$ for $h=\exp(-\zeta \delta)$.

Note that the $x$ variable only changes in ones of these steps so that $x_t = \hat{x}_t = \tilde{x}_{t+\delta}$, allowing us to eliminate the redundant terms. We can also eliminate $\hat{p}_t$ by substitution. We therefore need only to work in terms of the triple $(x_{t}, p_{t}, \tilde{p}_{t+\delta})$ at each combined time step. The conditional distribution for these remaining random variables is then:
\begin{equation}
B^{\theta}_k(x_{t}, p_{t}, \tilde{p}_{t+\delta}|x_{t+\delta},p_{t+\delta})=\delta_{\Phi^{-1}(x_{t+\delta},p_{t+\delta})}(x_t,\tilde{p}_{t+\delta})\mathcal{N}(p_t;h f_{\theta}(t+\delta,x_t,\tilde{p}_{t+\delta}),(1-h^2)M)    
\end{equation}

\noindent where $\delta_{\Phi^{-1}(x_{t+\delta},p_{t+\delta})}(x_t,\tilde{p}_{t+\delta})$ is a Dirac-delta centred on $\Phi^{-1}(x_{t+\delta},p_{t+\delta})$. From the main text we note that the forward Hamiltonian integrator has the conditional distribution:

\begin{equation}
F_{k+1}(\tilde{p}_{t+\delta},x_{t+\delta},p_{t+\delta}|x_{t},p_{t})=\delta_{\Phi(x_{t},\tilde{p}_{t+\delta})}(x_{t+\delta},p_{t+\delta})\mathcal{N}(\tilde{p}_{t+\delta};h p_t,(1-h^2)M).
\end{equation}

We again transition to discretized notation with $\delta:=T/K$, and $k=0,...,K$. Bringing together all the remaining random variables at each time step we have:
\begin{equation*}
 \textstyle Q(\textbf{x},\textbf{p})=\pi_0(x_0)\mathcal{N}(p_0;0,M)\prod_{k=0}^{K-1} F_{k+1}(x_{k+1},p_{k+1},\tilde{p}_{k+1}|x_k,p_k),
 \end{equation*}
 \begin{equation*}
 \textstyle \Gamma_\theta(\textbf{x},\textbf{p})=\gamma(x_K)\mathcal{N}(p_K;0,M)\prod_{k=0}^{K-1}B^{\theta}_{k}(x_{k},p_{k},\tilde{p}_{k+1}|x_{k+1},p_{k+1}).
 \end{equation*}
 
 We are interested in the unnormalized importance weight $w_\theta(\textbf{x},\textbf{p}) = \Gamma_\theta(\textbf{x},\textbf{p})/Q(\textbf{x},\textbf{p})$. We note that the division of delta functions is in general ill-defined and in our case should be interpreted formally in terms of the action of the volume preserving Leapfrog integrator flows $\phi_t$, as we describe in the main text. Further although the numerator and denominator in the definition of $w_\theta(\textbf{x},\textbf{p})$ do not have density with respect to Lebesgue measure, $\Gamma_\theta(\textbf{x},\textbf{p})$ has density with respect to $Q(\textbf{x},\textbf{p})$ and therefore $w_\theta(\textbf{x},\textbf{p})$ may be interpreted as the resulting Radon-Nikodym derivative. The explicit expression for $\log w_\theta(\textbf{x},\textbf{p})$ in equation (\ref{eq:AISOPTestimateHamiltonian}) of the main text then follows.
 
 We now show informally how maximizing the corresponding ELBO $\mathbb{E}_Q[\log w_\theta(\textbf{x},\textbf{p})]$ corresponds approximately to minimizing the score matching loss given in Proposition \ref{prop:ELBOunderdampedCT} for $\delta \ll 1$. We restrict the derivation to $M=I$ for simplicity. For pedagogical reasons, it is beneficial here to get back to the continuous time notations and recall that we use $k$ corresponds to $t_k=k\delta$. From direct calculations, maximizing the ELBO is equivalent to minimizing 
 \begin{align*}
J(\theta) & =\sum_{k=0}^{K-1}\mathbb{E}_{Q}\left[||p_{k\delta}-hf_{\theta}(k\delta,x_{k\delta},\widetilde{p}_{(k+1)\delta})||^{2}\right]
\end{align*}
where 
\begin{align*}
\mathbb{E}_{Q}\left[||p_{k\delta}-hf_{\theta}(k\delta,x_{k\delta},\widetilde{p}_{(k+1)\delta})||^{2}\right] & =\mathbb{E}_{Q}\left[||p_{k\delta}-h\left(\widetilde{p}_{(k+1)\delta}+2\delta\zeta[s_{\theta}(k\delta,x_{k\delta},\widetilde{p}_{(k+1)\delta})+\widetilde{p}_{(k+1)\delta}]\right)||^{2}\right]
\end{align*}
Note that $h=\exp(-\delta\zeta)\approx1-\delta\zeta$ so we obtain
\begin{align*}
&\mathbb{E}_{Q}\left[||p_{k\delta}-hf_{\theta}(k\delta,x_{k\delta},\widetilde{p}_{(k+1)\delta})||^{2}\right] \\
\approx & \mathbb{E}_{Q}\left[||p_{k\delta}-(1-\delta\zeta)\left(\widetilde{p}_{(k+1)\delta}+2\delta\zeta[s_{\theta}(k\delta,x_{k\delta},\widetilde{p}_{(k+1)\delta})+\widetilde{p}_{(k+1)\delta}]\right)||^{2}\right]\\
 \approx &\mathbb{E}_{Q}\left[||p_{k\delta}-(1+\delta\zeta)\widetilde{p}_{(k+1)\delta}-2\delta\zeta s_{\theta}(k\delta,x_{k\delta},\widetilde{p}_{(k+1)\delta})||^{2}\right]
\end{align*}
by neglecting terms of order $\delta^{2}$.
Now we try to further understand the asymptotic when $\delta \rightarrow 0$. We have that 
\begin{equation*}
(x_{k\delta},\widetilde{p}_{(k+1)\delta}) =\Phi_{k\delta}^{-1}(x_{(k+1)\delta},p_{(k+1)\delta}).
\end{equation*}
Now as we use for $\Phi_{k\delta}$ a leapfrog-type integrator, we do have \begin{equation*}
    \Phi_{k\delta}^{-1}(x',p')=\Phi_{\textup{flip}} \circ \Phi_{k\delta} \circ \Phi_{\textup{flip}}(x',p')
\end{equation*}
where $\Phi_{\textup{flip}}(x,p)=(x,-p)$ and, for $\delta \rightarrow 0$, we have
\begin{equation*}
    \Phi_{k\delta}(x,p) \approx ( x+\delta p, p- \delta \nabla E_{k \delta}(x)),
\end{equation*}
where $\pi_{k\delta}(x)\propto \exp(-E_{k\delta}(x))$. So we have for 
\begin{align*}
    \Phi_{k\delta}^{-1}(x',p')&=\Phi_{\textup{flip}} \circ \Phi_{t} \circ \Phi_{\textup{flip}}(x',p')\\
    &=\Phi_{\textup{flip}} \circ \Phi_{k\delta}(x',-p')\\
    &=\Phi_{\textup{flip}}(x'-\delta p',-p'-\delta \nabla E_{k\delta}(x'))\\
    &=(x'-\delta p',p'+\delta \nabla E_{k\delta}(x')).
\end{align*}
It follows that
\begin{align*}
&\mathbb{E}_{Q}\left[||p_{k\delta}-(1+\delta\zeta)\widetilde{p}_{(k+1)\delta}-2\delta\zeta s_{\theta}(k\delta,x_{k\delta},\widetilde{p}_{(k+1)\delta})||^{2}\right]\\
\approx & \mathbb{E}_{Q}\left[||p_{k\delta}-(1+\delta\zeta)(p_{(k+1)\delta}+\delta \nabla E_{k\delta}(x_{(k+1)\delta}))-2\delta\zeta s_{\theta}(k\delta,x_{(k+1)\delta},p_{(k+1)\delta})||^{2}\right]\\
\approx & \mathbb{E}_{Q}\left[||p_{k\delta}-p_{(k+1)\delta}-\delta(\zeta p_{(k+1)\delta}+\nabla E_{k\delta}(x_{(k+1)\delta})+2\zeta s_{\theta}(k\delta,x_{(k+1)\delta},p_{(k+1)\delta}))||^{2}\right].
\end{align*}
Now we have
\begin{align}
&\frac{1}{4\zeta\delta}\mathbb{E}_{Q}\left[||p_{k\delta}-p_{(k+1)\delta}-\delta(\zeta p_{(k+1)\delta}+\nabla E_{k\delta}(x_{(k+1)\delta})+2\zeta s_{\theta}(k\delta,x_{(k+1)\delta},p_{(k+1)\delta}))||^{2}\right] \nonumber\\
=&\delta \zeta \mathbb{E}_{Q}\left[||s_{\theta}(k\delta,x_{(k+1)\delta},p_{(k+1)\delta}) - \frac{1}{2\zeta \delta}(p_{k\delta}-p_{(k+1)\delta}-\delta(\zeta p_{(k+1)\delta}+\nabla E_{k\delta}(x_{(k+1)\delta})))||^{2}\right]\nonumber\\
\approx &\delta \zeta \mathbb{E}_{Q}\left[||s_{\theta}(k\delta,x_{(k+1)\delta},p_{(k+1)\delta}) - \frac{1}{2\zeta \delta}(p_{k\delta}-p_{(k+1)\delta}-\delta(\zeta p_{k\delta}+\nabla E_{k\delta}(x_{k\delta})))||^{2}\right]\nonumber\\
=&\delta \zeta \mathbb{E}_{Q}\left[||s_{\theta}(k\delta,x_{(k+1)\delta},p_{(k+1)\delta}) - \nabla_{p_{(k+1)\delta})} \log F(p_{(k+1)\delta}|p_{k \delta},x_{k \delta})||^{2}\right],\label{eq:approximateexpressionloss}
\end{align}
where $F(p_{(k+1)\delta}|p_{k \delta},x_{k \delta})=\mathcal{N}(p_{(k+1)\delta};(1-\delta \zeta)p_{k\delta}-\delta \nabla E_{k\delta}(x_{k\delta});2\zeta \delta I)$ and the joint distribution $F(x_{(k+1)\delta},p_{(k+1)\delta})|p_{k \delta},x_{k\delta})=\delta_{x_{k\delta}-\delta p_{k\delta}}(x_{(k+1)\delta})F(p_{(k+1)\delta}|p_{k \delta},x_{k \delta})$ is an Euler approximation of the forward transition of the underdamped Langevin dynamics. Now we expect similarly $\eta_{(k+1)\delta|k\delta}(x_{(k+1)\delta},p_{(k+1)\delta}|x_{k\delta},p_{k\delta})\approx \eta_{(k+1)\delta|k\delta}(p_{(k+1)\delta}|x_{k\delta},p_{k\delta})\eta_{(k+1)\delta|k\delta}(x_{(k+1)\delta}|x_{k\delta},p_{k\delta})$ for $\delta \ll 1$ and (\ref{eq:approximateexpressionloss}) is thus an approximation of the score matching loss.

\section{Diffusion processes: SGM, Langevin and AIS}\label{sec:sgmlangevinaisdiffusions}
We provide here a more detailed discussion between the similarities and differences between the diffusion process considered for SGM and the proposed approach. We first recall some basic elements of diffusion processes. Consider the diffusion $(x_t)_{t \in [0,T]}$  on $\mathbb{R}^d$
\begin{equation}\label{eq:generaldiffusion}
    \rmd x_t=f(t,x_t) \rmd t+\sqrt{2} \rmd B_t,\quad\quad x_0\sim q_0,
\end{equation}
where $(B_t)_{t\in [0,T]}$ is standard multivariate Brownian motion and $q_0$ is the initial distribution. 
The law $q_t$ of $x_t$ induced by this diffusion satisfies the Fokker--Planck--Kolmogorov equation 
\begin{equation}\label{eq:FPK}
    \frac{\partial  q_t(x)}{\partial t}=-\nabla \cdot [f(t,x) q_t(x)] +\Delta q_t(x) 
\end{equation}
where $\nabla \cdot [f(t,x) q_t(x)]=\sum_{i=1}^d \frac{\partial [f_i(t,x)q_t(x)]}{\partial x_i}$ and $\Delta q_t(x)=\sum_{i=1}^d \frac{\partial^2 q_t(x)}{\partial x_i^2}$ denote the divergence and Laplacian operators; see e.g. \cite{klebaner2012introduction}.

For SGM, we define the following Ornstein--Ulhenbeck process on $\mathbb{R}^d$ which corresponds to using 
\begin{equation}\label{eq:OU}
    f_{\textup{OU}}(t,x)=- x
\end{equation}
where $(B_t)_{t\in [0,T]}$ is standard multivariate Brownian motion and $q_0$ is the data distribution. This is also known in the SGM literature as the variance preserving diffusion \citep{song2020score}. This process adds noise progressively to the complex data distribution and converges geometrically fast to its invariant distribution which is the standard multivariate Gaussian $\pi_{\textup{OU}}(x)=\mathcal{N}(x;0,I)$ (see e.g. \citep{debortoli2021diffusion}) verifying indeed that the r.h.s. of (\ref{eq:FPK}) satisfies
\begin{align*}
    -\nabla \cdot [f_{\textup{OU}}(t,x) \pi_{\textup{OU}}(x)] + \Delta \pi_{\textup{OU}}(x)=0.
\end{align*}

More generally, a time-homogeneous Langevin diffusion to sample from a target distribution $\pi$ is of the form 
\begin{equation}\label{eq:OU}
    f_{\textup{Lgv}}(t,x)=\nabla \log \pi(x).
\end{equation}
It is easily check that $\pi$ is indeed an invariant distribution as the r.h.s. of (\ref{eq:FPK}) satisfies
\begin{align*}
    -\nabla \cdot [ f_{\textup{Lgv}}(t,x) \pi(x)] + \Delta \pi(x)=0,
\end{align*}
so that $q_t=\pi$ for all $t$ if $q_0=\pi$. Moreover, $q_t$ converges to $\pi$ whatever being $\pi_0$; see e.g. \cite{roberts1996exponential}. However, obtaining sharp quantitative bounds for complex $\pi$ is a more difficult task than for $\pi_{\textup{OU}}(x)=\mathcal{N}(x;0,I)$ in general.

In the context of this paper, the ``forward'' diffusion process we define is a \emph{time-inhomogeneous} Langevin algorithm
\begin{equation}\label{eq:OU}
    f_{\textup{AIS}}(t,x)=\nabla \log \pi_t(x),
\end{equation}
with $q_0=\pi_0$ an easy-to-sample distribution and $(\pi_t)_{t\in [0,T]}$ a non-constant curve of distributions such that $\pi_T=\pi$. So, contrary to SGM, which starts from a complex distribution and moves towards a simple distribution, we start here from a simple distribution and moves towards a complex distribution. 

In this scenario, even in $q_0=\pi_0$ then we do not have $q_t=\pi_t$ as the diffusion always lags behind its stationary distribution at time $t$. However, quantitative results measuring the discrepancy between the law of $x_T$ and $\pi$ for such annealed diffusions have been obtained; see e.g. \citep{grillo1994logarithmic,fournier2021simulated,tang2021simulated}. For this discrepancy to be small, one requires $\pi_t$ to vary slowly over time.

\section{Experimental Details}\label{sec:expe}

In all experiments, we sweep over diffusion time, number of steps, step-sizes and whether to learn them, and the annealing schedule. We identify the best parameters for each sampler individually on a validation set and then re-run these methods using 5 different seeds to obtain error bars on test set performance.
All experiments were executed on 8 GPUs for parallelized training and a single instance of our most expensive experiment (VAE) takes under 3 hours including evaluation.
Experiments are implemented in JAX~\citep{jax2018github} using the DeepMind JAX ecosystem~\citep{deepmind2020jax}.

\subsection{Sampler parameterization}
For all models, the step size was learned via a function $\epsilon_\theta(t)$ which is a 2-layer neural network with 32 hidden units, followed by a scaled sigmoid function which constrains $\epsilon_\theta(t) < 0.25$. As in prior work~\citep{Geffner:2021} we found this alleviated some instabilities in training. 

When learning the annealing schedule, we parameterize an increasing sequence of $T$ steps using unconstrained parameters $b_t$ (initialized to the same constant). We map these to our annealing schedule with
\begin{align}
    \beta_t = \frac{\sum_{t'\leq t}\sigma(b_{t'})}{\sum_{t'=1}^T \sigma(b_{t'})}
\end{align}
where we fix $\beta_0 = 0$ and $\sigma$ is the sigmoid function. This ensures that $\beta_0 = 0$, $\beta_K = 1$, and $\beta_t < \beta_{t'}$ when $t < t'$.

For UHA~\citep{Geffner:2021}, we also learn the momentum refreshment parameter $\eta \in (0, 1)$. We parameterize this with a parameter $u$ and define $\eta = .98 \sigma(u) + .01$ to keep the values in the range $(.01, .99)$ which we found alleviated sone training instabilities.

\subsection{Score model parameterization}
We parameterize our score model $s_\theta(t, x)$ using an MLP residual network. We first project the $x$ to dim $d_h$ using a linear layer and embed discrete time steps $t$ to dim $d_t$ using a learned embedding map. We then apply $k$ residual blocks.

Each block begins with a layer norm~\citep{ba2016layer} operation followed by a nonlinearity. We project the hidden representation to dim $2 \cdot d_h$ using a linear layer, project the embedding of $t$ to dim $2\cdot d_h$ using another linear map and add them together. We then apply another nonlinearity and then project the back to $d_h$ using another linear layer. We use the swish nonlinearity~\citep{ramachandran2017searching} throughout. 

To ensure our ELBO is initialized to a reasonable value we \emph{warm start} it so that at initialization, the score model outputs the standard AIS backward kernels. For the ULA version of our approach we do this by defining a score model $\tilde{s}_\theta(t, x)$ as explained above (but set the final layer weights to 0 at initialization) and define:
\begin{align}
s_\theta(t, x) = \tilde{s}_\theta(t, x) + \nabla_x \log \gamma_t(x).
\end{align}

For the UHA version we parameterize a score model $\tilde{s}_\theta(t, x, p)$ and define:
\begin{align}
s_\theta(t, x, p) = \tilde{s}_\theta(t, x, p) - M^{-1}p.
\end{align}

In both cases we found this led to much faster convergence and better results overall.

\subsection{Hyper-parameters}
In all experiments we use $k=3$ residual blocks in our score network. For our Gaussian experiments we set $d_h = 512$ and $d_t=16$. All models are trained with the Adam optimizer~\cite{kingma2014adam} with learning rates $0.001$ and $0.0001$ and up to 300k iterations with a batch size of 128. For static targets, we produce an estimate of $\log Z$ using 16,384 importance samples. As these methods produce a stochastic lower-bound on $\log Z$ we report the result from the hyper-parameter setting which gives the largest $\log Z$ estimate.

The VAE experiment uses architectures described in \cite{Burda2015}, which consists of encoder and decoder MLPs with two hidden layers with 200 units each, $\tanh$ activations, and 50 latent dimensions.
In contrast to \cite{Geffner:2021}, we found this architecture work better than the one described in \cite{Geffner:2021}, especially when trained for more iterations.
We chose the best performing models and their hyperparameters by monitoring validation performance during training.
We report performance of the best combination on the full test set, for each model respectively.
The best performance were reached with a matched the number of sampler steps between ULA/ULA-MCD and UHA/UHA-MCD -- 64 and 32 respectively.

\section{Additional Results}
\subsection{Static Targets}
\label{app:static}
Here we present additional results on a $\mathcal{N}(10, I)$ target, a $\mathcal{N}(0, 0.1 I)$ target, and a $\text{Laplace}(0, I)$ target. Results can be found in Tables \ref{tab:staticsgaussian}, \ref{tab:staticgaussstd}, and \ref{tab:staticlaplace}.

\begin{table}[h]
    \centering
    \begin{tabular}{c ||c|c||c|c || c  | c || c  | c }
    \hline
     Sampler  & \multicolumn{2}{c}{ULA}&  \multicolumn{2}{c}{UHA}& \multicolumn{2}{c}{ULA-MCD} & \multicolumn{2}{c}{UHA-MCD} \\
     $\#$ steps  &64&256  &64&256 &64&256  &64&256 \\
    \hline
    \hline
    Dim-20 & \makecell{-46.75 \\ $\pm$ 0.69} & \makecell{-6.23 \\ $\pm$ 0.91} & \makecell{0.0002 \\ $\pm$ 0.0008} & \makecell{0.0002 \\ $\pm$ 0.0004} & \makecell{-0.017 \\ $\pm$ 0.020} & \makecell{0.0034 \\ $\pm$ 0.0055} & \makecell{-0.0005 \\ $\pm$ 0.0007} & \makecell{0.0000 \\ $\pm$ 0.0005} \\
\hline
Dim-200 & \makecell{-752.25 \\ $\pm$ 2.43} & \makecell{-160.60 \\ $\pm$ 2.03} & \makecell{0.0003 \\ $\pm$ 0.0026} & \makecell{-0.0005 \\ $\pm$ 0.0007} & \makecell{-4.74 \\ $\pm$ 1.20} & \makecell{-0.019 \\ $\pm$ 0.047} & \makecell{0.0008 \\ $\pm$ 0.0032} & \makecell{0.0060 \\ $\pm$ 0.0084} \\
\hline
Dim-500 & \makecell{-1999.40 \\ $\pm$ 18.49} & \makecell{-455.20 \\ $\pm$ 7.90} & \makecell{0.0006 \\ $\pm$ 0.0007} & \makecell{-0.0008 \\ $\pm$ 0.0012} & \makecell{-21.62 \\ $\pm$ 1.64} & \makecell{-0.29 \\ $\pm$ 0.16} & \makecell{0.0030 \\ $\pm$ 0.0063} & \makecell{0.0099 \\ $\pm$ 0.0050} \\
\hline
\hline
    \hline
    \end{tabular}
    \caption{$\log Z$ estimates for a $\mathcal{N}(10, I)$ target. Averages and standard errors over 3 seeds.}
    \label{tab:staticsgaussian}
\end{table}

\begin{table}[h]
    \centering
    \begin{tabular}{c ||c|c||c|c || c  | c || c  | c }
    \hline
     Sampler  & \multicolumn{2}{c}{ULA}&  \multicolumn{2}{c}{UHA}& \multicolumn{2}{c}{ULA-MCD} & \multicolumn{2}{c}{UHA-MCD} \\
     $\#$ steps  &64&256  &64&256 &64&256  &64&256 \\
    \hline
    \hline
Dim-20 & \makecell{-1.58 \\ $\pm$ 0.42} & \makecell{-0.27 \\ $\pm$ 0.13} & \makecell{-4.46 \\ $\pm$ 0.93} & \makecell{-0.41 \\ $\pm$ 0.51} & \makecell{0.0095 \\ $\pm$ 0.0155} & \makecell{0.0057 \\ $\pm$ 0.0052} & \makecell{0.0038 \\ $\pm$ 0.0273} & \makecell{0.0002 \\ $\pm$ 0.0120} \\
\hline
Dim-200 & \makecell{-166.23 \\ $\pm$ 5.15} & \makecell{-58.33 \\ $\pm$ 1.99} & \makecell{-228.79 \\ $\pm$ 3.64} & \makecell{-74.62 \\ $\pm$ 1.60} & \makecell{-4.53 \\ $\pm$ 0.96} & \makecell{-0.67 \\ $\pm$ 0.33} & \makecell{-4.10 \\ $\pm$ 0.98} & \makecell{-1.47 \\ $\pm$ 0.57} \\
\hline
Dim-500 & \makecell{-545.77 \\ $\pm$ 5.55} & \makecell{-207.15 \\ $\pm$ 5.73} & \makecell{-704.96 \\ $\pm$ 7.30} & \makecell{-247.11 \\ $\pm$ 4.83} & \makecell{-27.36 \\ $\pm$ 2.23} & \makecell{-10.11 \\ $\pm$ 1.17} & \makecell{-29.20 \\ $\pm$ 3.46} & \makecell{-5.14 \\ $\pm$ 1.50} \\
\hline
    \hline
    \end{tabular}
    \caption{$\log Z$ estimates for a $\mathcal{N}(0, 0.1 I )$ target. Averages and standard errors over 3 seeds.}
    \label{tab:staticgaussstd}
\end{table}

\begin{table}[h]
    \centering
    \begin{tabular}{c ||c|c||c|c || c  | c || c  | c }
    \hline
     Sampler  & \multicolumn{2}{c}{ULA}&  \multicolumn{2}{c}{UHA}& \multicolumn{2}{c}{ULA-MCD} & \multicolumn{2}{c}{UHA-MCD} \\
     $\#$ steps  &64&256  &64&256 &64&256  &64&256 \\
    \hline
    \hline
    Dim-20 & \makecell{0.31 \\ $\pm$ 0.45} & \makecell{0.40 \\ $\pm$ 0.63} & \makecell{-0.0086 \\ $\pm$ 0.1314} & \makecell{-0.0077 \\ $\pm$ 0.1340} & \makecell{0.092 \\ $\pm$ 0.235} & \makecell{-0.23 \\ $\pm$ 0.02} & \makecell{0.0003 \\ $\pm$ 0.1573} & \makecell{-0.020 \\ $\pm$ 0.141} \\
\hline
Dim-200 & \makecell{-5.40 \\ $\pm$ 0.53} & \makecell{-4.54 \\ $\pm$ 0.96} & \makecell{-5.27 \\ $\pm$ 0.18} & \makecell{-5.28 \\ $\pm$ 0.19} & \makecell{-5.08 \\ $\pm$ 0.52} & \makecell{-4.31 \\ $\pm$ 0.66} & \makecell{-5.30 \\ $\pm$ 0.15} & \makecell{-5.39 \\ $\pm$ 0.09} \\

\hline
Dim-500 & \makecell{-17.67 \\ $\pm$ 1.91} & \makecell{-17.25 \\ $\pm$ 0.85} & \makecell{-17.91 \\ $\pm$ 0.78} & \makecell{-17.91 \\ $\pm$ 0.78} & \makecell{-17.64 \\ $\pm$ 1.98} & \makecell{-15.52 \\ $\pm$ 1.26} & \makecell{-18.11 \\ $\pm$ 0.70} & \makecell{-18.11 \\ $\pm$ 0.72} \\
\hline
    \hline
    \end{tabular}
    \caption{$\log Z$ estimates for a Student-T target. Averages and standard errors over 3 seeds.}
    \label{tab:staticlaplace}
\end{table}

\subsection{Normalizing Flow}
\label{app:flowtarg}
In this experiment we train NICE~\citep{dinh2014nice} flows which are fitted on downsampled variants of the MNIST dataset at resolutions $7 \times 7$, $14 \times 14$, and $28 \times 28$.
All models are trained for 100K steps with a batch size of 128 and then $\log Z$ is estimated using 4096 importance samples.

Results can be seen in Table \ref{tab:flow}.
In the largest setting we can see that UHA outperforms ULA, but our method outperforms both.
We note that a Sequential Monte Carlo sampler or AIS with Metropolis-Hastings corrections would be capable of estimating $\log Z$ very close to the true value of 0.
We have found that unadjusted samplers required for building a differentiable evidence lower-bound have difficulties sampling from this target distribution.
As discussed in the Limitations section of the main paper, we hypothesize that this is due to the fact that we are limited to using a relatively small number of transitions, since backpropagating through these repeated updates can be unstable.
We believe this is a limitation of this approach to inference and will impact any approach which utilizes an unadjusted forward sampler. Still, we note that our approach  to learning an optimized reversal (MCD) leads to improvements over the standard AIS reversals. We believe addressing these issues to be a key area for future research to focus.  

\begin{table}[h]
    \centering
    \begin{tabular}{c | c | c | c }
    \hline
       Dimension & ULA & UHA & MCD (ours) \\
    \hline
       $7 \times 7$   & 0.14 & 0.17  & \textbf{0.11} \\
       $14 \times 14$  &  13.24 & 15.04 & \textbf{6.25}  \\
       $28 \times 28$   & 141.29 & 82.16 & \textbf{23.10} \\
    \hline
    \end{tabular}
    \caption{$\log Z$ estimate absolute error for Normalizing flows.}
    \label{tab:flow}
\end{table}

\subsection{Score Network Ablation}
\label{app:network_ablation}
\begin{figure}[h]
    \centering
    \includegraphics[height=.35\textwidth]{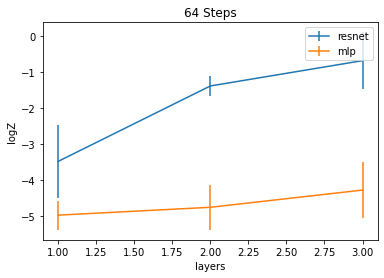}
    \includegraphics[height=.35\textwidth]{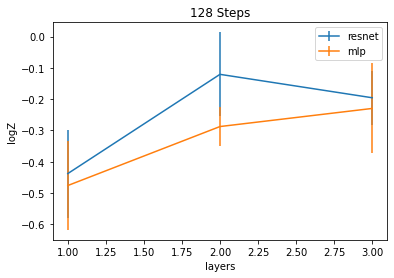}
    \caption{Score network architecture ablation on Gaussian mixture target (main results can be found in Table \ref{tab:staticmog}). Left: results with 64 sampling steps, right: results with 128 sampling steps. }
    \label{app:arch_rebuttal}
\end{figure}

We include additional results exploring the impact of various score-network architectures on performance. We re-run our Gaussian Mixture experiments in dimension 200 using 1) an MLP with residual connections and 2) a standard MLP, both with 1, 2, and 3 layers. We run these experiments with 64 and 128 sampling steps. Results can be seen in Figure \ref{app:arch_rebuttal}. We see that more expressive architectures lead to better performance in general. Further, we find that with more sampling steps, the impact of a more expressive score network diminishes. This aligns with intuition as, when using more steps we will use a smaller step size, and the standard AIS reversal becomes a better approximation to the true reversal.

\end{document}